\documentclass[times,colorlinks=true,allcolors=blue]{zHenriquesLab-StyleBioRxiv}

\usepackage{booktabs}
\usepackage{tabularx}
\usepackage{longtable}
\usepackage{pdflscape}
\usepackage{multirow}
\usepackage{xltabular}
\usepackage{orcidlink}
\usepackage{placeins}
\usepackage{doi}
\usepackage[labelfont={bf,sf},labelsep=period]{subcaption}

\leadauthor{X. Lekunberri}

\begin{document}

\title{Deep Learning for Accurate Vision-based Catch Composition in Tropical Tuna Purse Seiners}
\shorttitle{Deep Learning for Accurate Vision-based Catch Composition in Tropical Tuna Purse Seiners}

\author[1,\Letter]{Xabier Lekunberri\,\orcidlink{0000-0002-6766-9935}}
\author[1]{Ahmad Kamal\,\orcidlink{0009-0008-2135-3741}}
\author[2]{Izaro Goienetxea\,\orcidlink{0000-0002-1959-131X}}
\author[1]{Jon Ruiz\,\orcidlink{0000-0003-0717-0136}}
\author[1]{Iñaki Quincoces\,\orcidlink{0000-0003-4177-5497}}
\author[1,3]{Jaime Valls Miro\,\orcidlink{0000-0002-0083-7797}}
\author[2,3,4,5]{Ignacio Arganda-Carreras\,\orcidlink{0000-0003-0229-5722}}
\author[1]{Jose A. Fernandes-Salvador\,\orcidlink{0000-0003-4677-6077}}

\affil[1]{AZTI, Marine Research, Basque Research and Technology Alliance (BRTA), Txatxarramendi Ugartea z/g, Sukarrieta (Bizkaia), 48395, Spain}
\affil[2]{University of the Basque Country (UPV/EHU), San Sebastian, Spain}
\affil[3]{IKERBASQUE, Basque Foundation for Science, Bilbao, Spain}
\affil[4]{Donostia International Physics Center (DIPC), San Sebastian, Spain}
\affil[5]{Biofisika Institute (CSIC, UPV/EHU), Leioa, Spain}

\maketitle

\begin{abstract}
    Purse seiners play a crucial role in tuna fishing, as approximately 69\% of the world's tropical tuna is caught using this gear. All tuna Regional Fisheries Management Organizations have established minimum standards to use electronic monitoring (EM) in fisheries in addition to traditional observers. The EM systems produce a massive amount of video data that human analysts must process. Integrating artificial intelligence (AI) into their workflow can decrease that workload and improve the accuracy of the reports. However, species identification still poses significant challenges for AI, as achieving balanced performance across all species requires appropriate training data. Here, we quantify the difficulty experts face to distinguish bigeye tuna (BET, \textit{Thunnus Obesus}) from yellowfin tuna (YFT, \textit{Thunnus Albacares}) using images captured by EM systems. We found inter-expert agreements of 42.9\% $\pm$ 35.6\% for BET and 57.1\% $\pm$ 35.6\% for YFT. We then present a multi-stage pipeline to estimate the species composition of the catches using a reliable ground-truth dataset based on identifications made by observers on board. Three segmentation approaches are compared: Mask R-CNN, a combination of DINOv2 with SAM2, and a integration of YOLOv9 with SAM2. We found that the latest performs the best, with a validation mean average precision of 0.66 $\pm$ 0.03 and a recall of 0.88 $\pm$ 0.03. Segmented individuals are tracked using ByteTrack. For classification, we evaluate a standard multiclass classification model and a hierarchical approach, finding a superior generalization by the hierarchical. All our models were cross-validated during training and tested on fishing operations with fully known catch composition. Combining YOLOv9-SAM2 with the hierarchical classification produced the best estimations, with 84.8\% of the individuals being segmented and classified with a mean average error of 4.5\%.
\end{abstract}


\begin{corrauthor}
\href{mailto:xlekunberri@azti.es}{xlekunberri\at azti.es}
\end{corrauthor}

\section{Introduction}
\label{sec:tuna_2_introduction}

Tropical tuna species such as skipjack (SKJ, \textit{Katsuwonus pelamis}), yellowfin (YFT, \textit{Thunnus albacares}), and, to a lesser extent, bigeye (BET, \textit{Thunnus obesus}), are among the most heavily fished species worldwide \citep{fao_state_2024}. Purse seine stands out over other fisheries as it is responsible for approximately 69\% of the world's total catch of tropical tuna \citep{justel-rubio_snapshot_2024}. These three species play a significant role in pelagic ecosystems, acting as predators and prey, contributing to the balance and dynamics of marine food webs \citep{griffiths_just_2019}. Given the economic value and the key role these species play in their ecosystems, ensuring sustainable fishing with healthy stocks is crucial. In this regard, access to robust data and reliable statistics on catches is essential to ensure effective and sustainable fisheries management. The total weight of the catches made during a purse seine trip can be accurately estimated at landing, but the species composition remains problematic \citep{duparc_assessment_2018}. The catch composition is manually estimated during the fishing operations and recorded in the logbook. However, bias in logbooks has been evidenced since the beginning of the fishery, due to the difficulty in differentiating some species and the large number of individuals in each capture. This bias primarily concerns juvenile YFT and BET individuals, preventing the use of these records as precise estimators \citep{duparc_assessment_2020}.

Traditional fisheries monitoring implies that an onboard observer must supervise the fisher's actions directly at sea. The electronic monitoring (EM) systems, already in use in purse seiners, have been proposed as complement to onboard observers. These EM systems consist of several sensors installed onboard the fishing vessel, such as GPS receivers or CCTV cameras, and they are responsible for recording and storing data generated while the vessel is at sea \citep{mcelderry_at-sea_2008}. The EM can complement, or even replace, external observers, overcoming challenges such as the high cost and safety concerns of bringing additional personnel to the vessel \citep{fujita_designing_2018, van_helmond_electronic_2020}. However, manual review of EM footage is still tedious, as human analysts must watch hours of video and cross-reference it with sensor data to document fishing events. 

Several studies have evaluated the effectiveness and improvement over time of EM technology for catch quantification, species identification, and length/weight measurement \citep{briand_comparing_2018, briand_feasibility_2023, gilman_increasing_2019, monteagudo_preliminary_2014, murua_comparing_2020, murua_minimum_2022, ruiz_electronic_2015, ruiz_e-eye_2016}. These works revealed that EM footage analysis can accurately determine key metrics, such as the total tuna catch per set and fishing effort, but requires improvements regarding the accurate collection of species data. \citet{ruiz_comments_2021} demonstrated that, even if sampling is done onboard, obtaining precise estimates of species composition per fishing operation is often impractical without extensive and time-consuming sampling because of three main factors: (1) insufficient minimum sample size, (2) difficulty distinguishing juvenile YFT from BET, and (3) challenges in ensuring random sample selection. Such intensive sampling is incompatible with commercial operations, as minimizing the time between capture and freezing is crucial for preserving tuna quality \citep{duanmu_fast_2024}. Recent advances in artificial intelligence (AI) have led to the development of new methods that may overcome the challenges that traditional monitoring and EM still pose, providing a way to monitor the catches accurately. 

AI-based systems have already been developed in other fisheries to improve or facilitate species identification. For example, \citet{palmer_automatic_2022} estimated both the number and size of common dolphinfish (\textit{Coryphaena hippurus}) at a central fish auction. Using Mask R-CNN \citep{he_mask_2018} for segmentation, they achieved a mean average precision (mAP) of 79.8\% and a root mean squared error (RMSE) of 2.4 cm in length estimation. Other studies have opted for computationally faster architectures such as YOLO for detection and segmentation tasks. For instance, \citet{sokolova_integrated_2023} modified YOLOv5 to assign species and measure weights in discarded catches from demersal trawlers. They used a 5-fold cross-validation strategy, achieving a macro F1-score of 94.1\% and a weight prediction error of 29.74 grams. Building on this, \citet{cordova_multi-stage_2025} introduced a multi-stage approach that separated detection, classification, and weight estimation into distinct components, further improving accuracy.

Beyond task-specific models, one of the most significant advancements in recent years has been the emergence of foundation models \citep{bommasani_opportunities_2022}. Trained on massive, diverse datasets, these models provide general-purpose visual representations that can be adapted for a wide range of downstream applications. Despite their versatility, a recent survey indicates that foundation models still require domain-specific fine-tuning to achieve optimal performance \citep{jiaxing_sam2_2025}. Notable examples include DINOv2 \citep{oquab_dinov2_2024}, pretrained on a dataset of 142 million images for tasks such as depth estimation or semantic segmentation, and the Segment Anything Model 2 (SAM 2) \citep{ravi_sam_2024}, trained on over a billion masks for image and video segmentation. SAM2 enables zero-shot segmentation of arbitrary objects in photos or videos through simple prompts (e.g., points or bounding boxes). Early applications on underwater fish detection show promising results, but also reveal that their performance is highly dependent on the type of prompt \citep{kavasidis_training-free_2025, lian_evaluation_2024}.

Despite the remarkable accuracy in some studies, a critical gap remains between performance on controlled environments and the robustness required to implement the pipelines in real-world environments \citep{eickholt_advancing_2025}. In our previous work \citep{lekunberri_identification_2022} we achieved limited species identification in tuna purse seiners Using historical EM data. Several key points were identified that needed to be improved to develop a robust automated system: (1) the installation of a task-specific cameras providing a top-down view of the conveyor belt, (2) the availability of improved datasets for model training, and (3) accurate ground-truth data for operational testing, beyond statistical validation alone, to mitigate the impact of biased training samples.

This present work directly addresses these challenges by leveraging state-of-the-art deep learning methods to improve both segmentation and classification performance of our previous work. To our knowledge, this is the first study to process EM data collected onboard tuna purse seiners, where both train and test datasets derive from real fishing operations. This unique setting enables direct comparison of our computer vision-based pipeline against actual catch composition, granting a precise picture of the system’s performance.

\section{Material and methods}
\label{sec:tuna_2_material_and_methods}

\begin{figure*}[p]
	\centering
	\includegraphics[height=\textheight-4\baselineskip]{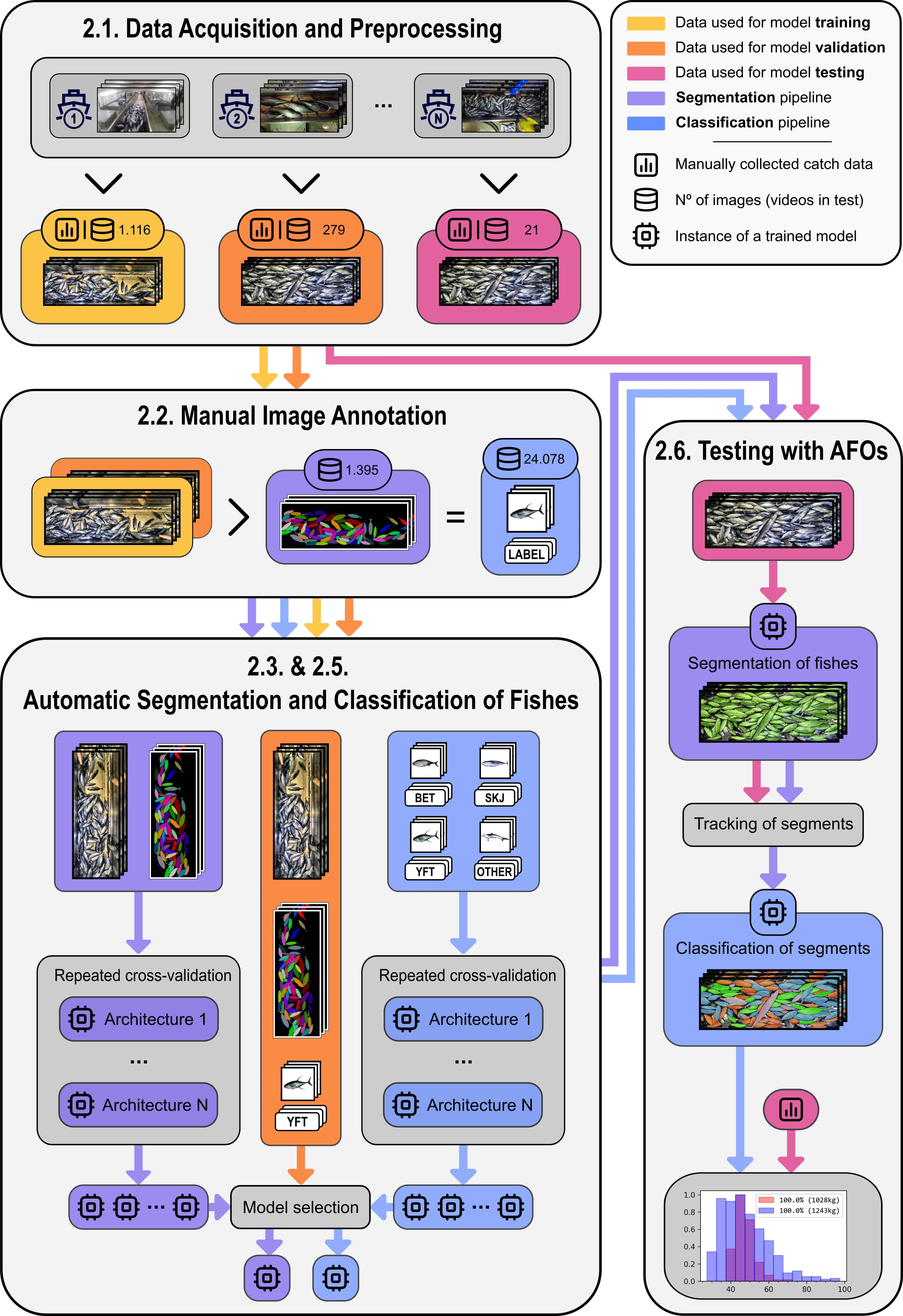}
	\caption{\textbf{Workflow that outlines the pipeline for fish image analysis.} It begins with the collection and manual annotation of images, followed by model training using cross-validation to select the best-performing architecture. The selected model is then used to automatically segment and classify fish in images. Finally, the pipeline is tested on real-world data. This process ensures robust and accurate fish identification.}
	\label{fig:tuna_2_1}
\end{figure*}

This section outlines the complete pipeline developed for estimating catches \autoref{fig:tuna_2_1}. The workflow consists of four main stages: (1) data acquisition, (2) image labeling, (3) model training and validation, and (4) ground-truth testing of the full pipeline.

Initial evaluation of historical EM footage was conducted in collaboration with nine taxonomy experts. The low agreement observed among experts regarding the identification of two of the three target species motivated the acquisition of new ground-truth data directly onboard tuna purse seiners. We selected videos in which the catch composition was perfectly known and only a single species appeared on the conveyor belt to train the models. On the other hand, videos with multiple species were used for the testing.

Training frames were extracted and uploaded to Computer Vision Annotation Tool (CVAT) for annotation \citep{cvat_ai_corporation_computer_2024}. Once annotated, segmentation and classification models were trained in parallel, as detailed in \autoref{sec:tuna_2_material_and_methods_automatic_segmentation} and \autoref{sec:tuna_2_material_and_methods_classification}, respectively. Two approaches were developed for both segmentation and classification, each trained and validated using repeated 5-fold cross-validation and subsequently tested on ground-truth data to simulate real-world performance. To assess the accuracy of the final predictions, results from the test phase were compared against the actual species distributions using a Wilcoxon signed-rank test.

\subsection{Data acquisition and preprocessing}
\label{sec:tuna_2_material_and_methods_data_acquisition}

Identifying species via images is not always feasible, as the traits that make species unique may not always be visible \citep{horton_recommendations_2021}. Two of the target species of this fishery, BET and YFT, are visually very similar, making it difficult or even impossible for analysts to identify them correctly. To quantify this difficulty, we conducted an experiment in which we brought together nine different experts in tuna identification to examine 40 different images. In that experiment, each one of the experts annotated all the tunas they could identify with a high degree of confidence. The dataset was restricted to BET and YFT individuals, since SKJ specimens are easy to identify. The low agreement among experts highlighted the difficulty of distinguishing BET from YFT in EM footage and led us to discard previously annotated data in favor of generating a new ground-truth dataset through an alternative approach. The findings will be discussed in the Results and Discussion sections.

The raw video data used in this work was recorded aboard vessels equipped with one or two conveyor belts in the fish processing area. Vessels with a single belt carried two EM cameras positioned near the bow and stern, while vessels with a second belt also included an additional camera on that belt., at the midpoint of the boat. All cameras recorded fishing operations at a 1920x1080 pixel resolution and 30 frames per second (FPS). During the trials at sea, our onboard personnel installed additional cameras to record extra footage. We used the ZED 2i camera for the first trip and upgraded to the ZED X series for the second. Both models are from Stereolabs and employ stereoscopic technology to capture the depth information in addition to standard RGB footage. Although 3D reconstruction was not explored in the present study, combining depth data with segmentation results would enable three-dimensional reconstruction of individual fish, potentially offering new opportunities for species classification as well as better size and weight estimation. 

The onboard scientists assisting with data collection had vast experience working on tuna purse seiners and were able to reliably identify target species in situ due to their background. Their knowledge was essential, as the data they collected was the basis for the pipeline described here.

During each fishing operation, the procedure was as follows: the conveyor belt was stopped, and all individuals from a given section were transferred into buckets. This process was repeated several times per operation. The number of individuals sampled varied depending on the total catch and the number of belt stoppages. Once the operation concluded, all sampled fish were individually identified to species, measured, and sorted to make up artificial fishing operations (AFOs) used for training and testing the models.

To do so, the identified tunas were placed back on the conveyor belt and recorded with the EM cameras under standard conditions. This procedure allowed us to simulate smaller-scale real fishing operations with fully known species composition and size distributions.  For training, the individuals were separated into batches by species before being returned to the conveyor belt, ensuring that all individuals selected for manual annotation were correctly labelled. In contrast, in AFOs used for testing purposes, the individuals were returned to the conveyor belt in a single mixed batch, emulating the conditions of real fishing operations.

Image preprocessing followed the methodology described in \citet{lekunberri_identification_2022}. The key step in this pipeline was the selection of the conveyor belt region through which the catches pass, correcting the perspective when necessary. This correction was not required when cameras were installed in a top-down position that fully covered the conveyor belt within the field of view and equipped with a lens that corrects optical distortions. Also, a CLAHE (Contrast Limited Adaptive Histogram Equalization) filter was applied to the images to homogenize the histograms and improve overall contrast. 

\subsection{Manual image annotation}
\label{sec:tuna_2_material_and_methods_manual_annotation}

All images were annotated using the Computer Vision Annotation Tool (CVAT), chosen for its maturity, team collaboration features, and the annotators’ prior experience with the software. Labels were organized hierarchically, with species-level annotations prioritized. Target species (BET, SKJ, and YFT) were labeled according to the 3-alpha code of the ASFIS List of Species for Fishery Statistics Purposes \citep{fao_asfis_2024}. 

In total, 1,395 images were annotated, yielding 24,078 individual fish segments. Some fishing operations from historical data were reused because of their exceptional quality and because the work of labeling them had already been done. In those cases, BET and YFT labels were changed for an additional joint label (BET\_OR\_YFT). That label and others generated by combining multiple categories were employed exclusively for the segmentation (\autoref{sec:tuna_2_material_and_methods_automatic_segmentation}) and the hierarchical classification approach (\autoref{sec:tuna_2_material_and_methods_classification}).

For non-target species captured as bycatch, all individuals were merged into a single NO\_TARGET label for classification, although species-level annotations were still collected to support future model improvements. A generic FISH label was added to maximize the number of annotated individuals for the segmentation, even though it did not carry species-specific information. The distribution of annotated segments across labels is shown in \autoref{tab:tuna_2_1}.

\begin{table*}[t]
	\centering
	\caption{\textbf{Number of segments used to train our models.} The segments are also broken down according to their label in a hierarchical manner. The top-level labels are TARGET and NO\_TARGET, corresponding to species targeted by this fishery (BET, SKJ, and YFT) and those not (all others). Target species are divided into SKJ and BET\_OR\_YFT depending on their visual similarity. Finally, BET\_OR\_YFT is divided into the corresponding species. The generic FISH label does not contain species-specific information.}
	\label{tab:tuna_2_1}
	\begin{tabularx}{\textwidth}{@{}>{\centering\arraybackslash}X 
	     >{\centering\arraybackslash}X 
	     >{\centering\arraybackslash}X 
	     >{\centering\arraybackslash}X 
	     >{\centering\arraybackslash}X@{}}
		\toprule
		\multicolumn{3}{c}{TARGET} & NO\_TARGET & FISH \\
		\multicolumn{3}{c}{18.232} & \multirow{5}{*}{1.204} & \multirow{5}{*}{4.642} \\ \cmidrule(r){1-3}
		SKJ & \multicolumn{2}{c}{BET\_OR\_YFT} & & \\ 
		\multirow{3}{*}{7.470} & \multicolumn{2}{c}{10.792} & & \\ \cmidrule(lr){2-3}
		& BET & YFT & & \\ 
		& 3.592 & 6.991 & & \\ \bottomrule
	\end{tabularx}
\end{table*}

\subsection{Automatic image segmentation}
\label{sec:tuna_2_material_and_methods_automatic_segmentation}

The conventional approach to segmentation and classification addresses both tasks in a single stage, assigning distinct labels to each species. In our case, however, we adopted a two-stage strategy. Since all target species share a highly similar morphology, we replaced the species-specific labels with a single class during the segmentation model's training. This approach allowed us to concentrate annotation efforts on a single segmentation class and to address classification separately using various techniques, as detailed in \autoref{sec:tuna_2_material_and_methods_classification}. 
The collected images and manual annotations (\autoref{fig:tuna_2_2}) were used to train and compare three segmentation approaches:

\begin{figure}[t]
	\centering
	\includegraphics[width=\columnwidth]{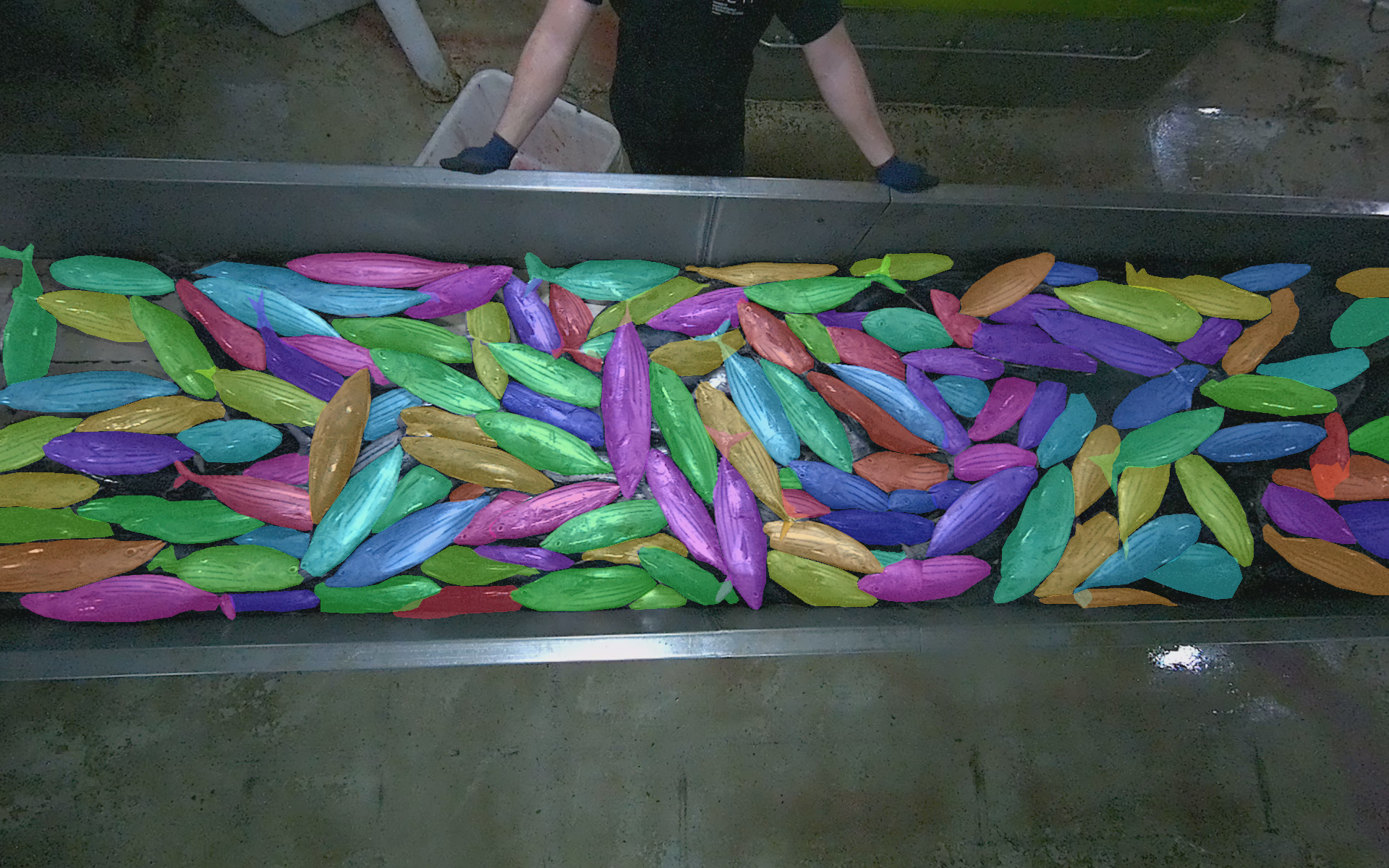}
	\caption{\textbf{Manually annotated image.} This image contains a masks for each fish on the conveyor belt that is visible from the point of view of the camera. This frame belongs to a monospecific SKJ sample. It was used during the training of the models.}
	\label{fig:tuna_2_2}
\end{figure}

\begin{enumerate}
    \item Mask R-CNN \citep{he_mask_2018},
    \item YOLOv9 \citep{wang_yolov9_2024} combined with SAM2 \citep{ravi_sam_2024}, and
    \item The pipeline described by \citet{gonzalez-marfil_dinosim_2025}, which performs zero-shot image segmentation using DINOv2 embeddings \citep{zhang_dino_2022} to compute similarity maps and employs SAM2 to generate segmentation masks of semantically similar objects. This last approach required no training or fine-tuning. 
\end{enumerate}

To expand the training set, we applied data augmentation. Image brightness, contrast, and saturation were randomly adjusted to simulate different lightning conditions. Random horizontal and vertical flips were applied to the images and masks to reduce orientation bias. \autoref{tab:tuna_2_supp_reproducibility} summarizes the computing infrastructure, hyperparameter search spaces, and data augmentation parameters used in model training.

Mask R-CNN, used in the first approach, is a two-stage instance segmentation algorithm that builds upon Faster R-CNN \citep{ren_faster_2017} by incorporating a branch for predicting segmentation masks on each Region of Interest (RoI). The process consists of two steps: the first generates proposals for regions that might contain objects, and the second refines these proposals and predicts masks. In our previous work \citep{lekunberri_identification_2022}, we used the “Mask R-CNN Inception ResNet V2 1024x1024” implementation from TensorFlow’s Object Detection API, from which we obtained a mean average precision (mAP) of 0.66. Since that API has been deprecated, we migrated to PyTorch and the Torchvision implementation.

In the second and third approaches, segmentation relied on SAM2, a state-of-the-art foundation model capable of segmenting images and videos across multiple domains. Without fine-tuning, SAM2 successfully segmented nearly every fish in our dataset. Its main limitation is the need for an initial prompt (bounding box or point) to locate a fish. The approaches differ in how these prompts were provided. In the second approach, we used the output of a fine-tuned YOLOv9 model as prompts for SAM2. In the third, we exploited DINOv2 embeddings to compute similarity maps based on a reference set of annotated fish, with SAM2 then generating masks for objects detected as semantically similar across other frames.

Segmentation models were evaluated through repeated cross-validation, with performance reported as the mean and standard deviation of both mAP and recall. A predicted segment was considered valid when its intersection over union (IoU) with the ground truth exceeded a predefined threshold. The IoU is measured as the area of the predicted bounding box that overlaps with the ground truth bounding box divided by the total area of both bounding boxes. The mAP is calculated as the area under the precision-recall curve across IoU thresholds from 0.05 to 0.95 in steps of 0.05.

\subsection{Individual fish tracking along the conveyor belt}
\label{sec:tuna_2_material_and_methods_tracking}

Since the fish can move freely on the conveyor belt and the belt occasionally stops, some individuals are captured in more frames than others. The videos are processed frame by frame, resulting in independent segmentations for each frame. Consequently, some individuals may be overrepresented, potentially biasing the final species estimations. Tracking each fish ensures that it is counted only once. 

In addition, having multiple images of the same fish might enhance the classification process, as it captures variations in lighting, pose and orientation across the camera's field of view. This variability improves the robustness of the classification by providing complementary visual information that would otherwise be omitted if only a single image were used. 

The original SAM2 implementation is capable of single-object tracking, and some extensions have added support for multiple-object tracking. However, we implemented our own tracking solution to ensure consistency across all segmentation approaches. Different algorithms for multi-object tracking, such as SORT \citep{bewley_simple_2017}, DeepSort \citep{wojke_simple_2017}, or ByteTrack \citep{zhang_bytetrack_2022}, were considered. Studies indicate that no single algorithm consistently outperforms the others, as a trade-off between computational efficiency and tracking precision must be considered \citep{abouelyazid_comparative_2023, alikhanov_online_2023, ferreira_dynamic_2024}.

We ultimately selected ByteTrack as the individuals in our images are moving in the same direction, and even if a segment's score was low due to partial occlusion, it was still used to try to match active trackers. Furthermore, ByteTrack’s high computational efficiency was also considered, as any delay introduced by tracking directly affects the overall processing time of the pipeline. We also included optical flow estimation to determine when operators stop or reverse the conveyor belt's direction. This allows us to pause tracking when the belt stops moving in the right direction and resume tracking when it starts moving again, which ensures that individuals that have left the camera’s field of view are not tracked when they reappear. 

\subsection{Fish classification by species}
\label{sec:tuna_2_material_and_methods_classification}

We evaluated two approaches for species classification:

\begin{enumerate}
    \item a standard multiclass classification, and 
    \item a hierarchical classification, where the final label is reached through multiple sequential decisions.
\end{enumerate}

In both approaches, we used the RegNetX400 architecture \citep{radosavovic_designing_2020}. As in the segmentation step, random brightness, contrast, and saturation modifications were used as data augmentation techniques, in addition to random horizontal and vertical flips. See \autoref{tab:tuna_2_supp_reproducibility} for the computing infrastructure, hyperparameter search spaces, and data augmentation parameters used during training.

The first approach follows a conventional multiclass scheme similar to our previous work \citep{lekunberri_identification_2022}, where a pretrained ResNet50V2 model from TensorFlow’s Object Detection API was used. That model achieved approximately 75\% of correct classifications. Since the API was deprecated, we migrated to PyTorch and adopted a newer architecture for improved performance and long-term support.

The second approach follows a hierarchical classification strategy, in which labels are grouped in pairs and progressively refined. At each level, one class is final (requiring no further classification), while the other proceeds to the next decision level. This hierarchy includes three classification stages: 

\begin{enumerate}
    \item Non-target species vs target species.
    \item Skipjack tuna vs other tunas.
    \item Bigeye tuna vs yellowfin tuna. 
\end{enumerate}

This structure was designed to increase classification complexity gradually. Although hierarchical classification requires multiple inferences per sample—thus increasing computational time—it enables finer discrimination between species. Similar hierarchical strategies have been successfully applied to tuna classification in simpler contexts, where individuals do not overlap and are photographed separately \citep{mujtaba_hierarchical_2022}. In both approaches, multiple predictions of the same fish were averaged to produce a final class prediction. As the same individual appears in several frames under varying lighting conditions and orientations, this aggregation increases robustness and reduces sensitivity to visual variability.

Finally, all classification strategies were tested independently and in combination. The final system may integrate any subset of these methods, depending on their performance trade-offs. Although hierarchical and weighted schemes introduce a computational cost, they can substantially improve classification accuracy, making the trade-off advantageous in practice.

\subsection{Statistical validation and ground truth testing}
\label{sec:tuna_2_material_and_methods_validation_and_gt}

All models were validated using robust, stratified, and repeated cross-validation. Except for YOLOv9, which was limited to five repetitions due to training time constraints, all segmentation and classification models underwent 10-times 5-fold repeated cross-validation. Model performance was evaluated using task-appropriate metrics: segmentation models were assessed with \textbf{standard COCO metrics}, including recall and mean average precision (mAP) across multiple segment sizes and intersection over union (IoU) thresholds, while classification models were evaluated through confusion matrices and accuracy measures. After validation, all models were tested on out-of-sample data. We used only the portion of the AFOs collected during sea trials that had not been used for training. These samples were collected exclusively for this work, ensuring that ground truth and modeled data corresponded to the same individuals.

The AFOs were processed through the entire pipeline (segmentation, tracking, and classification) with every possible approach combination to assess the overall performance. Finally, Wilcoxon signed-rank tests were employed to determine if statistically significant differences existed between the actual species distribution and the predictions generated by our pipeline.

\section{Results}
\label{sec:tuna_2_results}

This section presents the outcomes of: (1) the expert-based experiment assessing the ability to distinguish between BET and YFT using EM images, (2) the training and validation of all models, and (3) the comparison of model predictions against the ground-truth data collected onboard different vessels. 

\subsection{BET and YFT classification based only on images}
\label{sec:tuna_2_results_experts}

Nine experts in tuna taxonomy independently identified BET and YFT individuals in a set of 40 EM images. Experts were instructed to label only those fish they could identify with high confidence. Three experts identified substantially more tunas than the others, resulting in a total of 257 unique individuals being annotated (\autoref{fig:tuna_2_3}). Refer to \autoref{tab:tuna_2_supp_experts} in the Supplementary Materials to view all the labels.

\begin{figure}[t]
	\begin{subfigure}[t]{\columnwidth}
		\centering
		\includegraphics[width=\columnwidth]{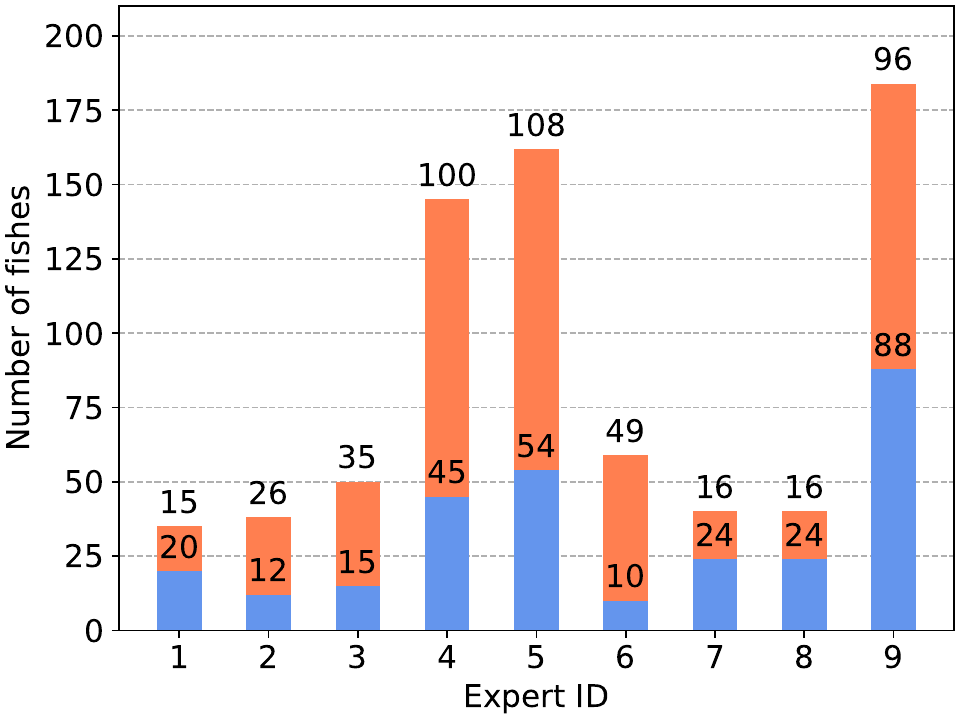}
		\caption{Number of individuals annotated by each of the nine experts.}
		\label{fig:tuna_2_3_a}
	\end{subfigure}
	\par\bigskip
	\begin{subfigure}[t]{\columnwidth}
		\centering
		\includegraphics[width=\columnwidth]{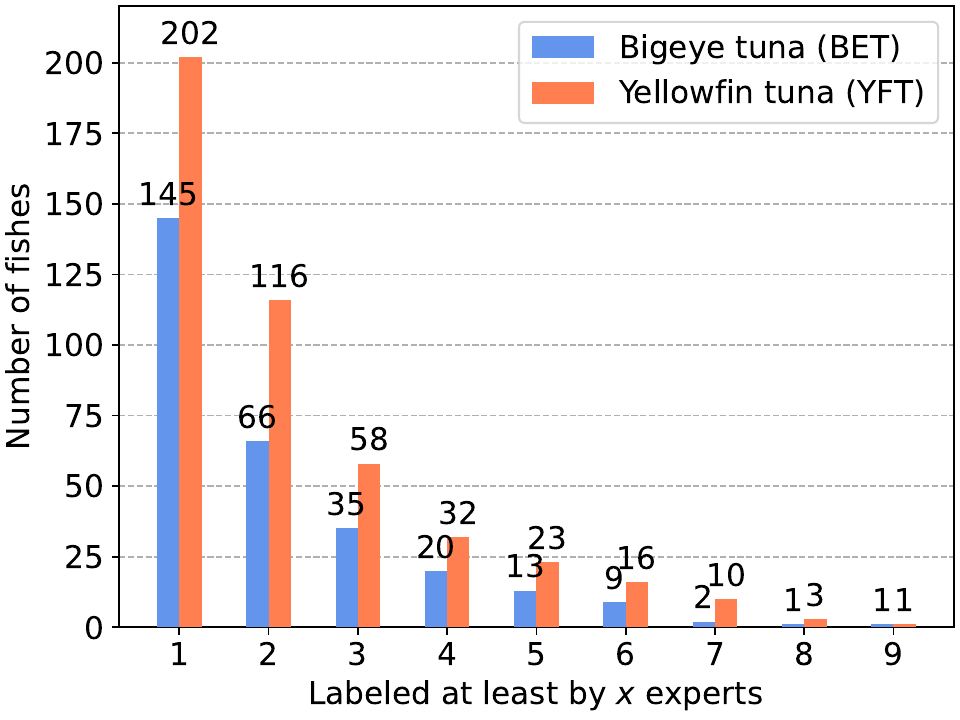}
		\caption{Number of individuals, split by species, that have been labeled at least by X experts.}
		\label{fig:tuna_2_3_b}
	\end{subfigure}
	\par\bigskip
	\caption{\textbf{Different statistics for the analysis based on expert identification.}}
	\label{fig:tuna_2_3}
\end{figure}

To calculate the inter-expert agreements, individuals identified by at least four (more than a third) experts were selected. We found that the agreement rate was 42.9\% $\pm$ 35.6\% for BET and 57.1\% $\pm$ 35.6\% for YFT. There was only consensus among the nine experts for one individual per species. These results indicate that experts cannot agree on identifying BET and YFT when only electronic monitoring images are available. Consequently, we concluded that remote annotation could not provide sufficient ground truth accuracy. To overcome this limitation, new data were acquired \textbf{in situ} onboard tuna purse seiners, where experts could visually confirm the species of each fish during the fishing operation. This ensured that the training footage for our models contained only one species at a time, eliminating ambiguity in subsequent image annotation.

\subsection{Segmentation of the images}
\label{sec:tuna_2_results_segmentation}

The combination of YOLOv9 and SAM2 yielded the best performance in terms of both mAP and recall, with statistically significant improvements on the test AFOs, as will be shown later. The cross-validated Mask R-CNN achieved a mAP of 0.63 $\pm$ 0.01 for bounding boxes and 0.67 $\pm$ 0.01 for segmentation masks. YOLOv9, on the other hand, obtained a mAP of 0.66 $\pm$ 0.03. In terms of recall, Mask R-CNN reached 0.69 $\pm$ 0.01, while YOLOv9 achieved 0.85 $\pm$ 0.03. 

\begin{figure}[t]
	\begin{subfigure}[t]{.49\columnwidth}
		\centering
		\includegraphics[width=\columnwidth]{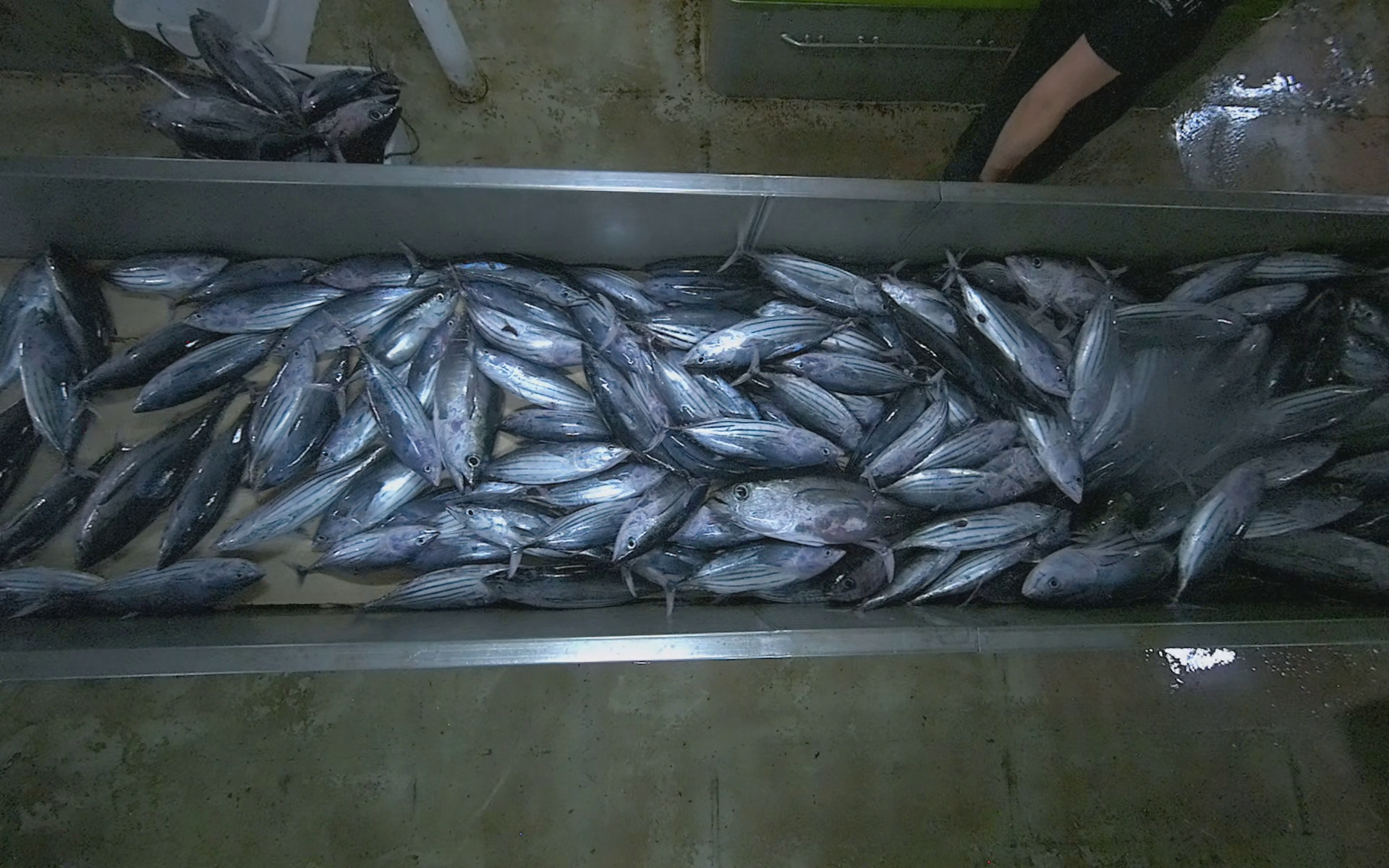}
		\caption{Original frame.}
		\label{fig:tuna_3_4_a}
	\end{subfigure}
	\hfill
	\begin{subfigure}[t]{.49\columnwidth}
		\centering
		\includegraphics[width=\columnwidth]{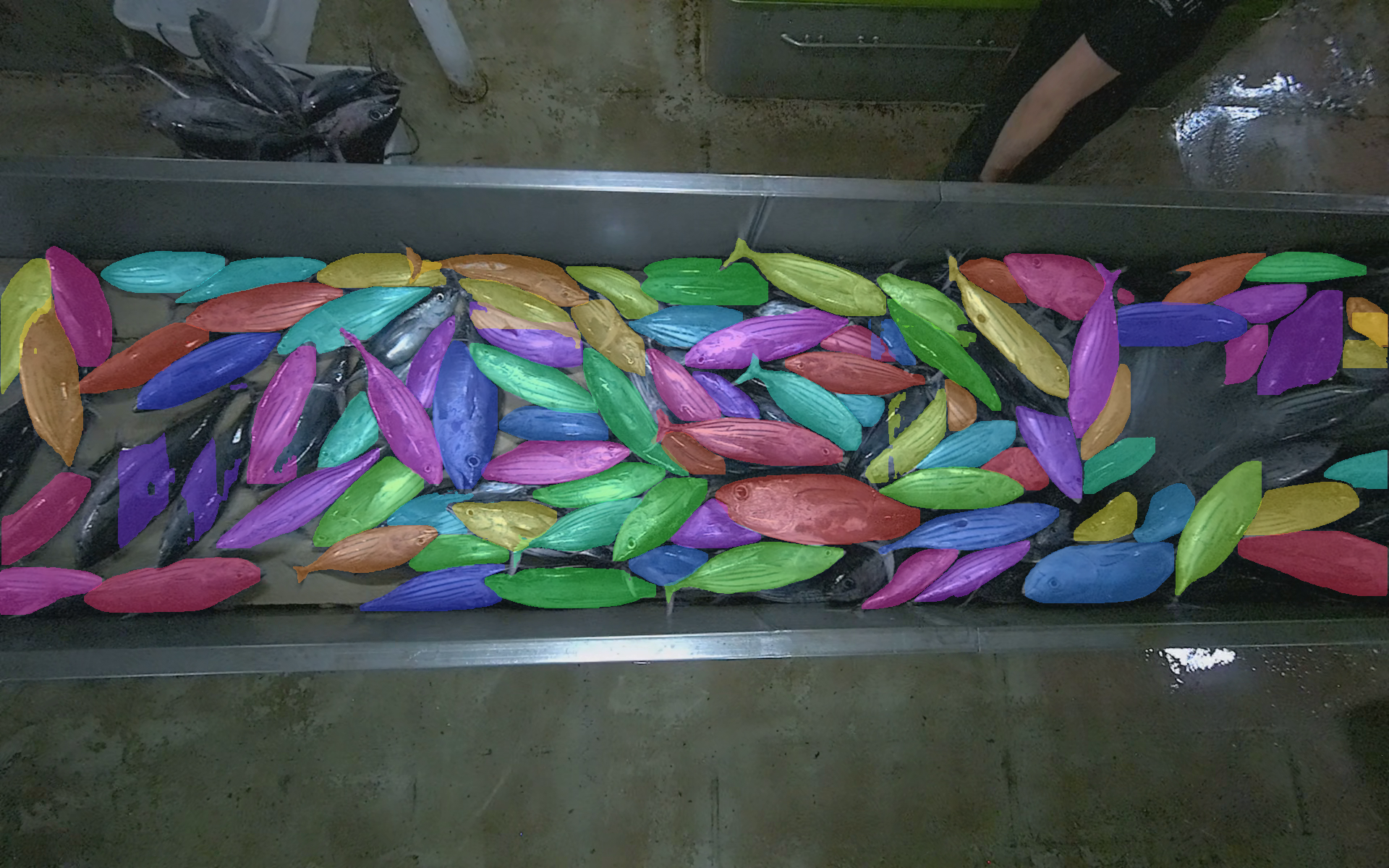}
		\caption{Segmented using Mask R-CNN.}
		\label{fig:tuna_2_4_b}
	\end{subfigure}
	\par\bigskip
	\begin{subfigure}[t]{.49\columnwidth}
		\centering
		\includegraphics[width=\columnwidth]{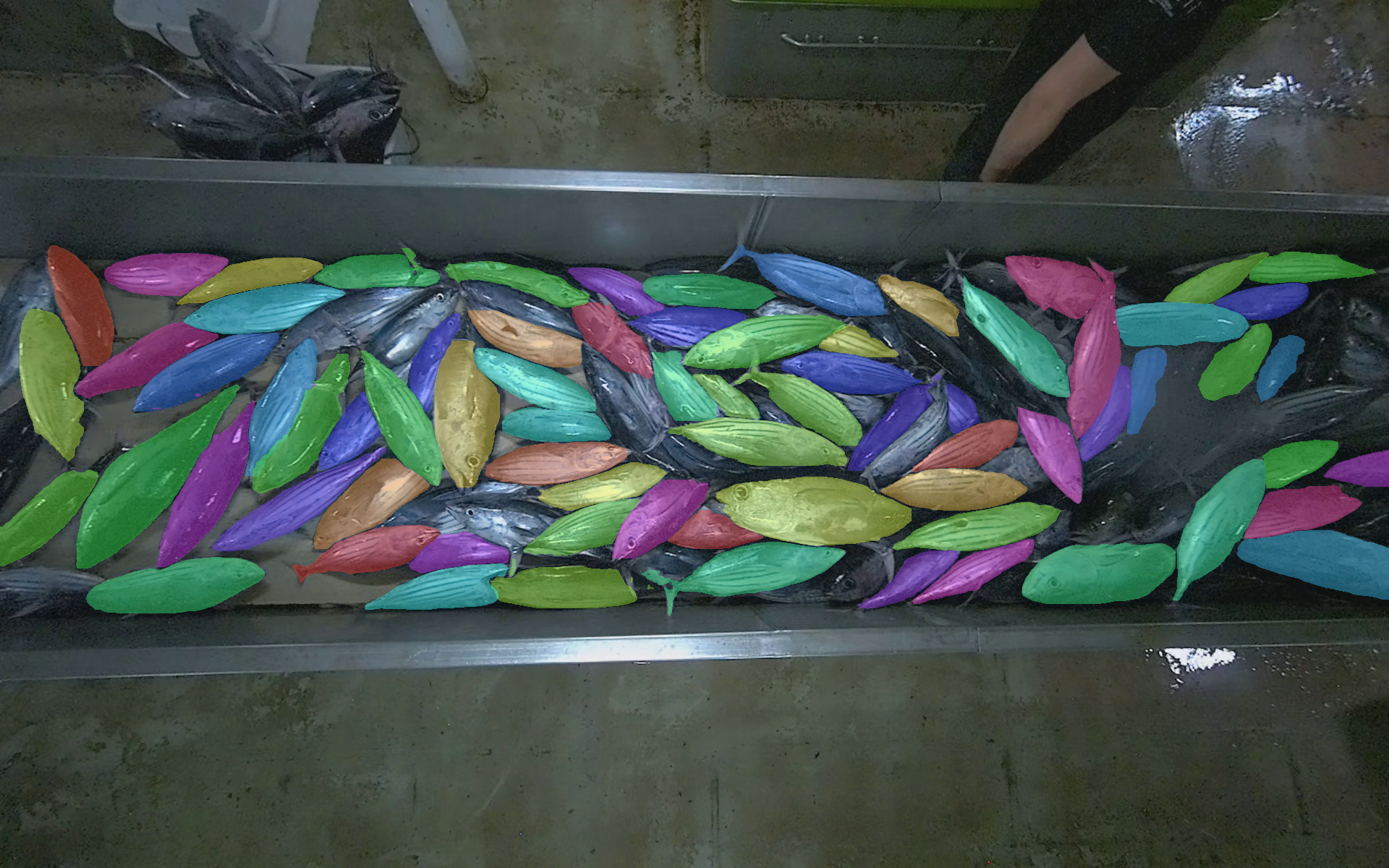}
		\caption{Segmented using YOLOv9 + SAM2.}
		\label{fig:tuna_2_4_c}
	\end{subfigure}
	\hfill
	\begin{subfigure}[t]{.49\columnwidth}
		\centering
		\includegraphics[width=\columnwidth]{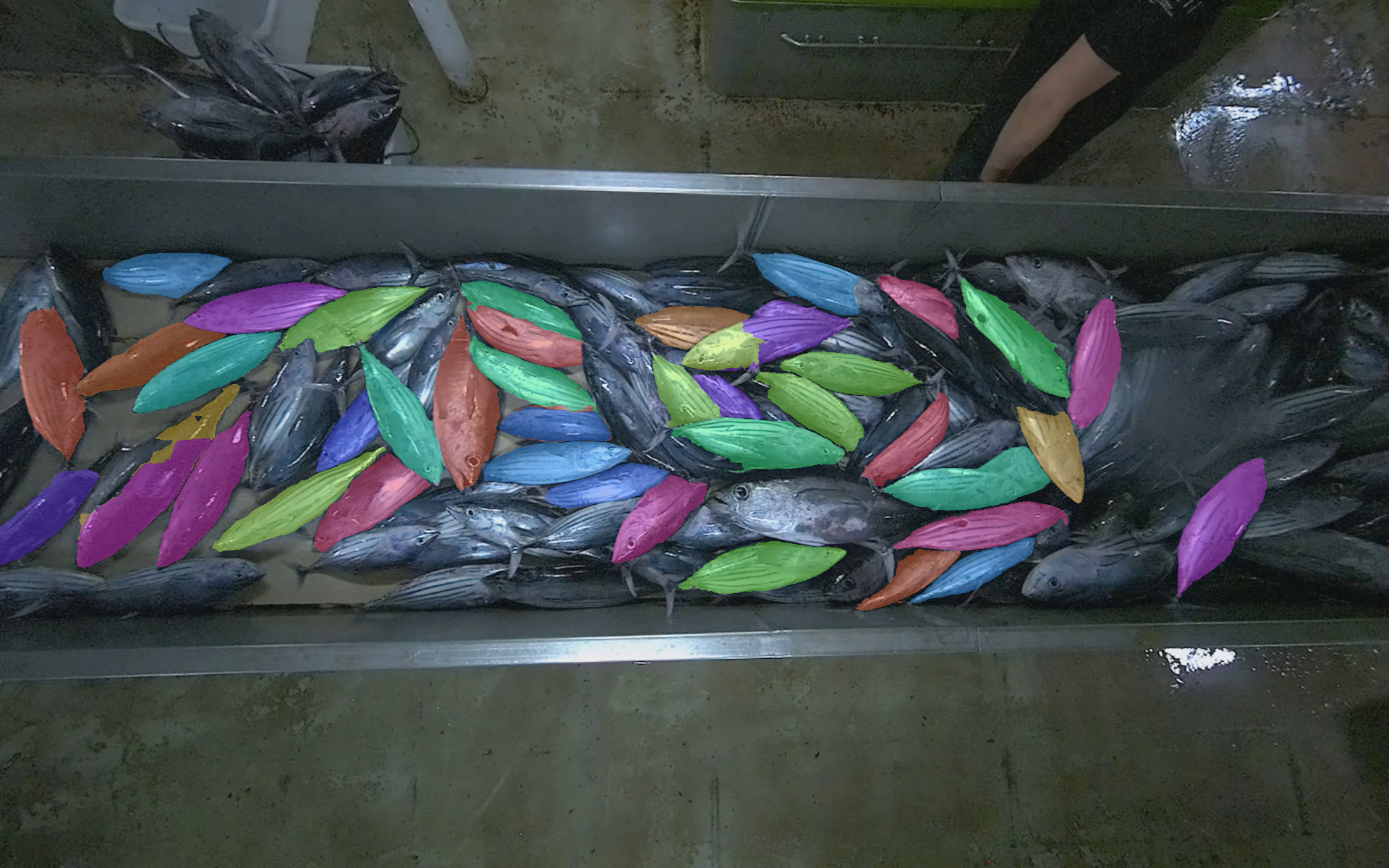}
		\caption{Segmented using DINOv2 + SAM2.}
		\label{fig:tuna_3_4_d}
	\end{subfigure}
	\par\bigskip
	\caption{\textbf{Comparison of all the segmentation approaches.} It can be noted that the right part of the camera has a small water droplet, making the fish in that area blurry and more difficult to segment.}
	\label{fig:tuna_2_4}
\end{figure}

No segmentation scores are reported for the second approach, as SAM2 weights were used without fine-tuning. Similarly, we do not provide segmentation scores for the third approach, where DINO was used to compute similarity maps, since this method also does not involve model fine-tuning. \autoref{fig:tuna_2_4} shows a raw frame from one of the AFOs together with the corresponding segmentations obtained by each approach.
Following the method described by \citet{gonzalez-marfil_dinosim_2025}, we chose not to include this method in further comparisons, as generating a single frame took up to ten minutes on a laptop GPU. Integrating such a time-consuming process into a future onboard application running on an embedded device would not be feasible, as near-real-time performance is required. 

\subsection{Classification of the fishes by species}
\label{sec:tuna_2_results_classification}

\begin{figure}[t]
	\centering
	\begin{subfigure}[t]{0.62\columnwidth}
		\centering
		\includegraphics[width=\columnwidth]{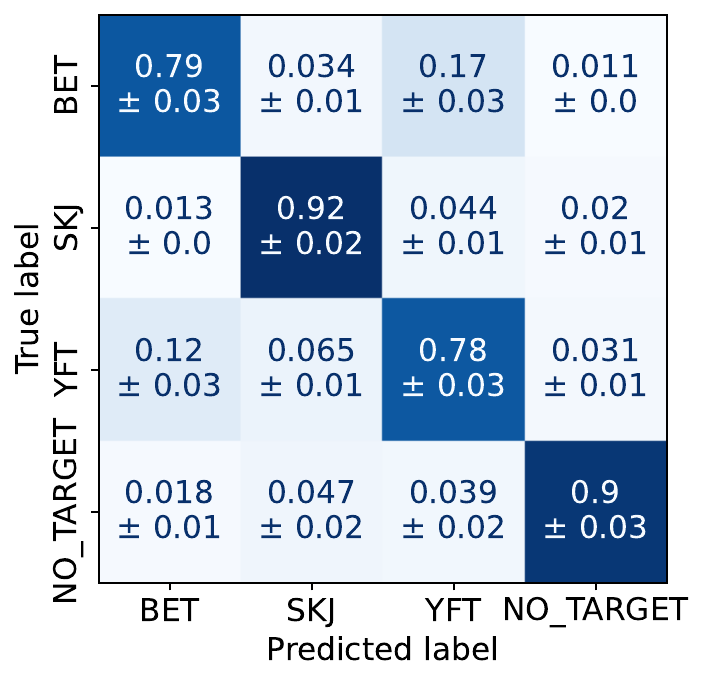}
		\caption{Standard classification.}
		\label{fig:tuna_3_5_a}
	\end{subfigure}
	\par\bigskip
	\begin{subfigure}[t]{0.62\columnwidth}
		\centering
		\includegraphics[width=\columnwidth]{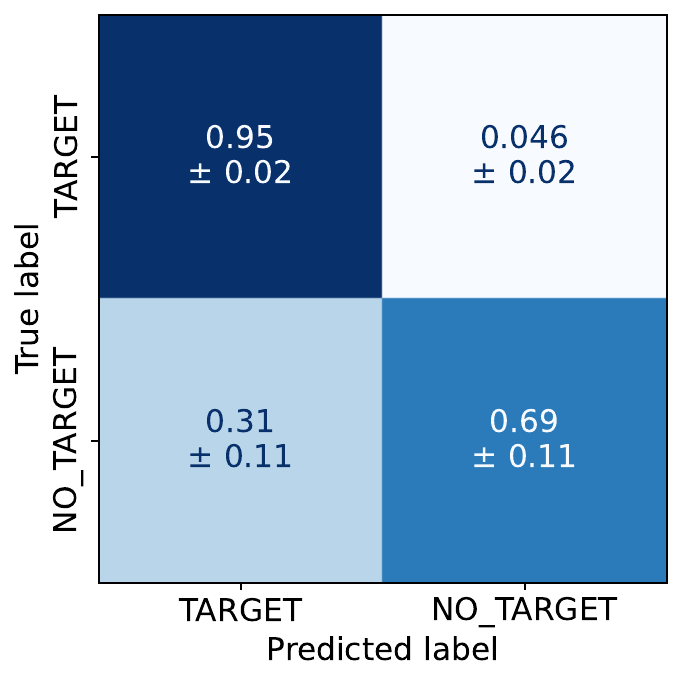}
		\caption{Hierarchical classification, step 1/3.}
		\label{fig:tuna_2_5_b}
	\end{subfigure}
	\par\bigskip
	\begin{subfigure}[t]{0.62\columnwidth}
		\centering
		\includegraphics[width=\columnwidth]{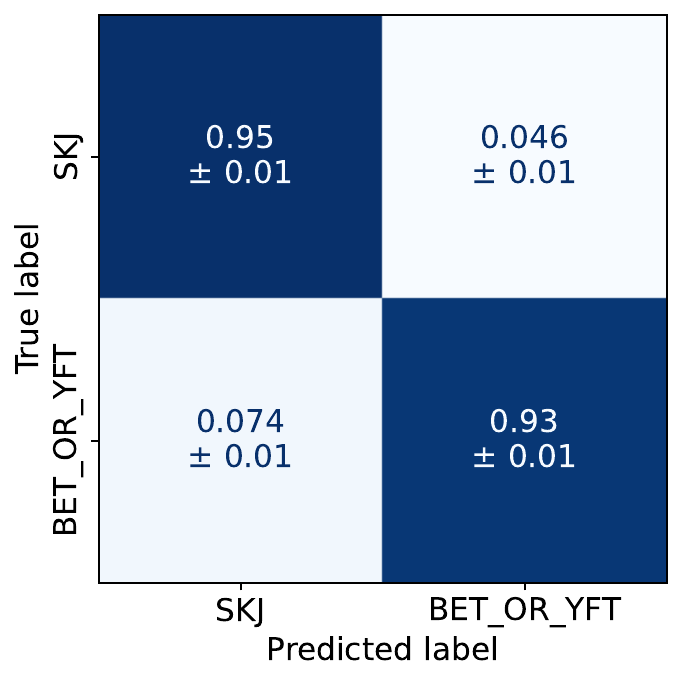}
		\caption{Hierarchical classification, step 2/3.}
		\label{fig:tuna_2_5_c}
	\end{subfigure}
	\par\bigskip
	\begin{subfigure}[t]{0.62\columnwidth}
		\centering
		\includegraphics[width=\columnwidth]{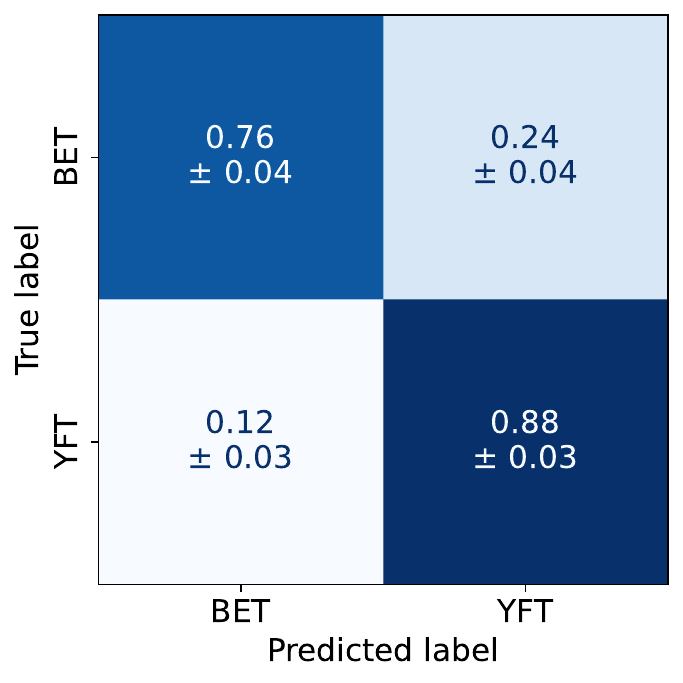}
		\caption{Hierarchical classification, step 3/3.}
		\label{fig:tuna_3_5_d}
	\end{subfigure}
	\par\bigskip
	\caption{\textbf{Confusion matrices for the four classification models.} These matrices are built using the validation data.}
	\label{fig:tuna_2_5}
\end{figure}

Four classification models were trained: one using a standard multiclass approach and three using a hierarchical scheme. As with segmentation, all models were cross-validated and tested on out-of-sample data. \autoref{fig:tuna_2_5} shows the confusion matrices and corresponding mean accuracies. 

Both approaches achieved comparable overall validation scores, but the main difference lies in the NO\_TARGET class, where the non-hierarchical classifier obtained a significantly better accuracy (0.90 $\pm$ 0.03 vs 0.69 $\pm$ 0.11). This may be explained by (i) the heterogeneous nature of the NO\_TARGET label, which groups multiple non-targeted species, including tuna-like fish visually similar to target tunas, and (ii) the limited number of training images for these classes. This imbalance is much more noticeable in the hierarchical classifier, as a one-vs-rest strategy binarizes each step. It should be noted that bycatch species were not targeted during the data gathering step, and that the label has not been prioritized in the development of this work. 

For the targeted species, SKJ showed consistently high accuracy in both approaches (0.92 $\pm$ 0.02 and 0.95 $\pm$ 0.01), while BET and YFT exhibited the lowest performance, as expected given their visual similarity. Even within the hierarchical framework, notable confusion persisted. Approximately one out of every four BET specimens is misclassified as yellowfin tuna YFT. Although in the hierarchical model the error is somewhat higher for this class, the proportion of bycatch in the total catch rarely exceeds 5\%. Therefore, in absolute terms of catch, the error is not very significant. 

\begin{table*}[t]
	\centering
	\caption{\textbf{Mean and the standard deviation of the percentage of the catches segmented by each segmentation approach.} Several Wilcoxon signed-rank tests were run for \textit{p} \textless{} 0.10 (*), \textit{p} \textless{} 0.05 (\textdagger), and \textit{p} \textless{} 0.01 (\textdaggerdbl), with the null hypothesis being that there are no differences in the amount of individuals segmented by the two approaches.}
	\label{tab:tuna_2_3}
	\begin{tabularx}{\textwidth}{@{}XXXX@{}}
		\toprule
		& First trip (\textdaggerdbl) & Second trip (\textdaggerdbl) & Both trips (\textdaggerdbl) \\ \midrule
		Mask R-CNN & 46.4\% $\pm$ 6.4\% & 63.5\% $\pm$ 11.5\% & 52.1\% $\pm$ 11.7\% \\
		YOLOv9 + SAM2 & 60.3\% $\pm$ 9.6\% & 84.8\% $\pm$ 13.5\% & 68.5\% $\pm$ 16.0\% \\ \bottomrule
	\end{tabularx}
\end{table*}

\begin{table*}[t]
    \centering
    \caption{\textbf{Mean average error (MAE) for classifying individuals segmented by the best approach (i.e., YOLOv9 + SAM2).} The best classification approach (i.e., standard or hierarchical) was selected for each trip. Several Wilcoxon signed-rank tests were run for \textit{p} \textless{} 0.10 (*), \textit{p} \textless{} 0.05 (\textdagger), and \textit{p} \textless{} 0.01 (\textdaggerdbl) with the null hypothesis being that there are no differences in the fish counted by the two methods (i.e., the ground truth and the evaluated one).}
    \label{tab:tuna_2_4}
    \begin{tabularx}{\textwidth}{@{}XXXX@{}}
        \toprule
        & First trip   & Second trip & Both trips   \\ \midrule
        BET  & 2.8\% $\pm$ 2.4\%  & 2.1\% $\pm$ 2.1\% & 2.3\% $\pm$ 2.3\% (*) \\
        SKJ  & 22.9\% $\pm$ 11.4\% (\textdaggerdbl) & 5.0\% $\pm$ 4.7\% & 19.8\% $\pm$ 11.0\% (*) \\
        YFT  & 20.7\% $\pm$ 10.8\% (\textdaggerdbl) & 5.6\% $\pm$ 8.0\% & 16.1\% $\pm$ 11.7\% (\textdaggerdbl) \\
        NO\_TARGET & 2.8\% $\pm$ 1.8\% (\textdaggerdbl) & 5.4\% $\pm$ 2.8\% & 3.8\% $\pm$ 4.5\%  \\ \bottomrule
    \end{tabularx}
\end{table*}

\subsection{Testing with artificial fishing operations}
\label{sec:tuna_2_results_testing_afo}

We evaluated the complete pipeline using 21 artificial fishing operations (AFOs) containing multiple species, from two fishing trips conducted on the same vessel (14 and 7 AFOs, respectively). Both segmentation methods segmented more than half of the caught tunas, but YOLOv9 + SAM2 clearly outperformed Mask R-CNN (68.5\% vs 52.1\%; \autoref{tab:tuna_2_3}). This superiority was consistent across trips (60.3\% vs 46.4\% and 84.8\% vs 63.5\%) and statistically significant in all cases (Wilcoxon signed-rank, \textit{p} < 0.01). Therefore, we selected YOLOv9 + SAM2 as the preferred segmentation approach.

Using this segmentation, we compared the standard and hierarchical classification schemes. Here, we calculate the error for a given species as the absolute difference between the predicted abundance and the ground truth. The standard classifier achieved a lower mean absolute error (MAE) overall (10.5\% vs 17.8\%), corresponding to roughly two-thirds of the hierarchical model’s error. The results varied by trip. The standard approach performed better on data collected during the first trip (12.3\% vs 24.5\%), while the hierarchical classification slightly improved results on data collected from the second trip (4.5\% vs 6.9\%). The differences between the ground truth and the predictions of our model also differ, as the errors for the second trip are much smaller for both standard and hierarchical approaches. Species-wise, using our best segmentation and classification approaches, we got a MAE of 2.1\% $\pm$ 2.1\% for BET, 5.0\% $\pm$ 4.7\% for SKJ, 5.6\% $\pm$ 8.0\% for YFT, and 5.4\% $\pm$ 2.8\% for NO\_TARGET (\autoref{tab:tuna_2_4}). Distribution tests confirmed significant differences only for the first trip. Detailed per-set results are provided in \autoref{tab:tuna_2_supp_mask} and \autoref{tab:tuna_2_supp_yolo_sam} of the Supplementary Material.

\section{Discussion}
\label{sec:tuna_2_discussion}

This work advances the current state of the art in automatic catch identification for tuna purse seiners by generating a reliable, ad hoc ground truth dataset, testing more effective segmentation and classification methods, and validating results against independent ground truth datasets. Comparative analysis identified YOLOv9 + SAM2 as the most effective segmentation approach, while a hierarchical binary classification strategy enhanced species discrimination. Tracking individual fish along the conveyor belt further stabilized predictions by averaging outputs across frames. 

Data quality has been critical to achieving these results, as expert-labeled datasets underpin every stage of the pipeline. However, species identification from EM imagery remains challenging. Key diagnostic traits are not always visible \citep{horton_recommendations_2021}, and even expert judgements can vary widely. The issue of observer bias and reliability in visual identification has also been extensively studied in citizen science contexts. Platforms like \textit{iNaturalist} rely on large-scale, non-expert labeling, and several studies have shown that accuracy strongly depends on participants’ familiarity with the species \citep{ackland_method_2024, mesaglio_expert_2025}. Although expert identifications are generally expected to be more reliable, evidence shows they can also be inconsistent. \citet{culverhouse_experts_2003} reported self-consistency as low as 67\% and agreement between experts of only 43\% when identifying plankton species. Our experiment revealed similar variability: inter-expert agreements of 42.9\% $\pm$ 35.6\% for BET and 57.1\% $\pm$ 35.6\% for YFT were found. Dropping to only one individual of each species being identified by all nine experts as the same species. We concluded that the agreements were not high enough to rely on those identifications as our ground truth. A posterior analysis by \citet{culverhouse_human_2007,austen_species_2018} analyzed human factors that may play a role in species identification. Both conclude that expert accuracy in identification is often negatively affected by boredom and fatigue, regardless of their experience. Given the repetitive nature of their job, these two conditions can be very prevalent among workers whose job is to review EM footage. This does not imply experts cannot distinguish these species, but rather that EM imagery alone is insufficient for reliable identification. In our study, onboard experts could handle the fish directly and confirm diagnostic traits, making our ground truth—and therefore our models—more robust. Although our pipeline’s performance is not perfect, its transparent and unbiased design increases confidence for both authorities and industry users.

The rapid shift in AI research towards large-scale foundation models \citep{schneider_foundation_2024} also affects fisheries monitoring. We evaluated DINOv2 and SAM2, following the approach of \citet{gonzalez-marfil_dinosim_2025}, in which DINOv2 is used to calculate similarity maps in the image and SAM2 to generate segmentation masks. While this zero-shot method showed promise---segmenting multiple tunas from a few user clicks as seen in \autoref{fig:tuna_2_4}---it was computationally unfeasible, taking up to ten minutes per frame. Moreover, such approaches still require validated ground truth. As foundation models and hardware continue to improve, real-time applications of this kind may become viable. AI-driven monitoring pipelines are increasingly common across marine research, from aquaculture optimization to ecosystem surveys. Public datasets such as Fishnet \citep{kay_fishnet_2021} and FDWE \citep{sokolova_integrated_2023} have accelerated these developments. \citet{cordova_multi-stage_2025, vanessen_automatic_2021, xu_rsnc-yolo_2024} are some prominent works in that field where even better validation scores than ours are obtained, but given that different species and environments are processed, a direct one-to-one comparison is not possible. Our results, based on a custom-made dataset, achieved a validation mAP of 0.66 $\pm$ 0.03 and a recall of 0.85 $\pm$ 0.03, an improvement over our previous iteration, segmenting nearly all visible tunas, whilst acknowledging that no model can segment individuals that are not visible in the lower layers of the conveyor belt. To achieve a higher proportion of visible and segmented fish, improvements will depend less on the model itself and more on the way fish are handled and loaded into the wells of these vessels, that is, on fish handling practices. Classification accuracy also improved substantially, 20\% for SKJ and 12\% for YFT. On the other hand, validation accuracy dropped by 6\% for BET. It should be noted that the total number of individuals for both training and validation has increased significantly in this study. Specifically, BET has a 32-fold increase (from 122 to 3,592 individuals), which makes the results much more robust. Nonetheless, standard deep learning metrics should not be the sole evaluation criteria; as \citet{eickholt_advancing_2025} emphasized, performance must be assessed in relation to management objectives and real-world usability. Our tests, conducted under near-operational conditions, provide robust validation despite the limited number of sampled fish.

Results differed between the two fishing trips used for testing. Both followed identical procedures, but the first used a rolling shutter camera and the second a global shutter model. The latter avoids motion-induced distortions caused by fast-moving fish, resulting in a dramatic reduction in MAE—from 12.3\% to 4.5\%—. Statistical analyses confirmed significant differences only for the rolling shutter data. Thus, global shutter cameras clearly enhance accuracy and should be preferred in future deployments.

Although these results represent a significant step forward, several aspects could further improve the pipeline’s operational deployment. One promising line is integrating depth (RGB-D) information. Our 3D cameras provide depth maps aligned with RGB images, enabling direct estimation of fish size and even weight through volumetric reconstruction \citep{ruchay_live_2022}. Similar approaches have improved semantic segmentation \citep{wang_brief_2021} and could enhance species-level classification, as morphometric traits differ among species \citep{iccat_iccat_2006}. Incorporating environmental variables that can be used to forecast fishing grounds \citep{goikoetxea_machine-learning_2024} could also help distinguish visually similar species like BET and YFT, which occupy different habitats \citep{arrizabalaga_global_2015}. 

While this study focuses on tropical tuna purse seiners, the pipeline is adaptable to other fisheries. For instance, minimal fine-tuning could make it suitable for albacore tuna (\textit{Thunnus Alalunga}) fisheries in the Cantabrian Sea, where catches are homogeneous and individually retrieved. In contrast, more heterogeneous fisheries, such as demersal bottom trawling, would require major adaptations due to their diverse species composition and debris-filled imagery.

Finally, we plan to generate additional AFOs to improve model robustness, particularly for minority species like BET and bycatch. As fewer individuals are included in these AFOs, error rates on less common species may be overestimated. Although our current seven validated operations demonstrate that the approach works, a even bigger testing effort is essential to warranty that a system like that could be applied to the whole fishery. Nevertheless, the present results clearly show that we are on the right track.

\section{Conclusions}
\label{sec:tuna_2_conclusions}

National administrations and Regional Fisheries Management Organizations (RFMOs) are increasingly adopting standards for the implementation of EM systems. In the specific case of tuna RFMOs, all have recently adopted these standards, which has encouraged a growing number of fishing companies to choose this option to comply with monitoring requirements. This is leading to a substantial increase in the volume of data that needs to be analyzed. In the specific context addressed here, there is currently no effective method for accurately estimating species-specific catch data from the tropical purse seine fleet. Traditional approaches, such as onboard observer sampling or visual interpretation of EM-derived imagery, have yielded imprecise estimates. This novel methodology has the potential to fill that gap, providing more reliable data that are critical for effective fisheries management, including stock assessment and regulatory enforcement. Although further work is needed, the progress made can be summarized as follows:

\begin{enumerate}
    \item Given the similarity between both species, experts disagree on identifying bigeye tuna (BET, \textit{Thunnus Obesus}) and yellowfin tuna (YFT, \textit{Thunnus Albacares}) when only EM footage is available. The observed agreement among experts was 42.9\% $\pm$ 35.6\% for BET and 57.1\% $\pm$ 35.6\% for YFT, which we considered insufficient for ground-truth annotation. This variability highlights the need to account for observer bias when training and validating AI systems for species classification.
    \item SAM2 can segment the catches of this fishery without requiring fine-tuning. When prompted with YOLOv9’s bounding boxes, the combined approach achieved near-complete segmentation of all fish visible to the camera. 
    \item While standard multiclass and hierarchical classification models yielded similar validation metrics, hierarchical classification demonstrated superior performance in real fishing operations. This finding supports the idea that the evaluation of AI systems should extend beyond statistical validation and involve real-world performance evaluation.
    \item Camera characteristics strongly influence accuracy. In tuna purse seiners, the rapid motion of fish on the conveyor belt causes motion blur when using rolling-shutter sensors, leading to a MAE of 12.3\% between our predictions and the actual species composition. Switching to global-shutter sensors reduced this error to 4.5\%.
    \item If a vessel's EM system is equipped with a camera with a global shutter sensor and captures images in a top-down perspective, the pipeline developed in this work can accurately estimate the composition of the catches with a MAE of 4.2\% $\pm$ 4.9\% for target species (BET, SKJ, and YFT) and 5.4\% $\pm$ 2.8\% for the bycatch species.
\end{enumerate}

\begin{acknowledgements}
    Xabier Lekunberri has received a PhD grant from the IKERTALENT Programme of the Department of Food, Rural Development, Agriculture, and Fisheries of the Basque Government. Open Access funding provided by the University of the Basque Country (EHU). We want to thank the skippers and crew members of the vessels that facilitated our data collection, the observers who accompanied us during both trips (Arnaitz Mugerza and Gorka Ocio), and the observers who worked tirelessly to annotate the images used to train the models (Arnaitz Mugerza, Lluis Horrach, Xuban Zumeta, and Iñaki Oyarzabal). The authors would also like to thank all the experts who generously participated in the classification test of yellowfin tuna (YFT) and bigeye tuna (BET) based on EM images. This paper is contribution no. XXX from AZTI, Marine Research, Basque Research and Technology Alliance (BRTA). The authors used DeepL Write for language improvement during manuscript preparation; all scientific content and interpretations are the authors’ own.  This work is partially supported by grant GIU23/022 funded by the University of the Basque Country (UPV/EHU), and grant PID2021-126701OB-I00, funded by the Ministerio de Ciencia, Innovación y Universidades, AEI, MCIN/AEI/10.13039/501100011033, and by "ERDF A way of making Europe”. This work has been funded by the European Union through the Horizon Europe program projects EVERYFISH (Grant agreement number 101059892) and OptiFish (Grant Agreement number 101136674) as well as BioBoost+ project within the Biodiversa+ European Biodiversity Partnership program. Views and opinions expressed are however those of the authors only and not necessarily reflect those of the European Union or Fundación Biodiversidad.
\end{acknowledgements}

\appendix

\section*{Bibliography}
\phantomsection
\addcontentsline{toc}{section}{Bibliography}
\bibliography{tuna_2}

\renewcommand{\thesection}{Supplementary Material~\Alph{section}}
\renewcommand{\thetable}{\Alph{section}.\arabic{table}}
\renewcommand{\thefigure}{\Alph{section}.\arabic{figure}}

\onecolumn
\section{Reproducibility}

\begin{table}[h]
    \caption{Summary of computational resources, data augmentation parameters, hyperparameter configurations, and training durations used for segmentation and classification models.}
    \label{tab:tuna_2_supp_reproducibility}
    \begin{subtable}{\columnwidth}
        \caption{Computing infrastructure used to train the models.}
        \label{tab:tuna_2_supp_reproducibility_comp}
        \begin{tabularx}{\textwidth}{@{}XXX@{}}
            \toprule
            Component & Model & Quantity \\ \midrule
            CPU & Intel\textsuperscript{\textregistered} Xeon\textsuperscript{\textregistered} Silver 4509Y & 2 \\
            GPU & NVIDIA\textsuperscript{\textregistered} H100 NVL (94GB) & 1 \\
            RAM & --- & 768GiB \\ \bottomrule
        \end{tabularx}
    \end{subtable}
    \par\vspace{0.5cm}
    \begin{subtable}{\columnwidth}
        \caption{Data augmentation used to train segmentation and classification models. V\textsubscript{B,C,S} represent the image's original brightness, contrast, and saturation values, respectively.}
        \label{tab:tuna_2_supp_reproducibility_data_aug}
        \begin{tabularx}{\textwidth}{@{}XXX@{}}
            \toprule
            Parameter & Possible values & Probability \\ \midrule
            Brightness & {[}max(0, 1 \textminus V\textsubscript{B}), 1 + V\textsubscript{B}{]} & 1 \\
            Contrast & {[}max(0, 1 \textminus V\textsubscript{C}), 1 + V\textsubscript{C}{]} & 1 \\
            Saturation & {[}max(0, 1 \textminus V\textsubscript{S}), 1 + V\textsubscript{S}{]} & 1 \\
            Horizontal flip & --- & 0.5 \\
            Vertical flip & --- & 0.5 \\ \bottomrule
        \end{tabularx}
    \end{subtable}
    \par\vspace{0.5cm}
    \begin{subtable}{\columnwidth}
        \caption{Values assigned to the hyperparmeters of the trained \textbf{segmentation models}.}
        \label{tab:tuna_2_supp_reproducibility_hyper_segm}
        \begin{tabularx}{\textwidth}{@{}XXX@{}}
            \toprule
            Hyperparameter & Search space & Best assignment \\ \midrule
            \multirow{2}{*}{Epochs} & Mask R-CNN [30, 40, 50, 60] & 40 \\
             & YOLOv9 [500, 600, 700, 800, 900] & 700 \\
            Batch size & {[}2, 4{]} & 2 \\
            Train / validation / test & {[}0.7 / 0.2 / 0.1{]} & 0.7 / 0.2 / 0.1 \\
            Optimizer & {[}SGD{]} & SGD \\
            Learning Rate & {[}0.01{]} & 0.01 \\
            Momentum & {[}0.95{]} & 0.95 \\
            Weight Decay & {[}0.0005{]} & 0.0005 \\
            LR Scheduler & {[}Exponential(gamma=0.96){]} & Exponential(gamma=0.96) \\ \bottomrule
        \end{tabularx}
    \end{subtable}
    \par\vspace{0.5cm}
    \begin{subtable}{\columnwidth}
        \caption{Values assigned to the hyperparmeters of the trained \textbf{classification models}.}
        \label{tab:tuna_2_supp_reproducibility_hyper_class}
        \begin{tabularx}{\textwidth}{@{}XXX@{}}
            \toprule
            Hyperparameter & Search space & Best assignment \\ \midrule
            Epochs & [40, 50, 60] & 40 \\
            Batch size & [2, 4] & 4 \\
            Train / validation / test & [0.7 / 0.2 / 0.1] & 0.7 / 0.2 / 0.1 \\
            Optimizer & [SGD] & SGD \\
            Learning Rate & [0.01] & 0.01 \\
            Momentum & [0.95] & 0.95 \\
            Weight Decay & [0.0005] & 0.0005 \\
            LR Scheduler & [StepLR(step\_size=7, gamma=0.1)] & StepLR(step\_size=7, gamma=0.1) \\ \bottomrule
        \end{tabularx}
    \end{subtable}
    \par\vspace{0.5cm}
    \begin{subtable}{\columnwidth}
        \caption{Time required to train all the models. This includes all folds and repetitions of the repeated cross-validation. In the hierarchical model, the three levels are added together.}
        \label{tab:tuna_2_supp_reproducibility_time}
        \begin{tabularx}{\textwidth}{@{}XXX@{}}
            \toprule
            \multicolumn{2}{l}{Model} & Train duration \\ \midrule
            \multirow{2}{*}{Segmentation} & Mask R-CNN & 21 days \\
             & YOLOv9 & 25 days \\
            \multirow{2}{*}{Classification} & Standard & 10 days \\
             & Hierarchical & 21 days \\ \bottomrule
        \end{tabularx}
    \end{subtable}
\end{table}

\newcolumntype{R}{>{\raggedleft\arraybackslash}X}
\begin{landscape}
	\vspace*{\fill} 
	\section{Mask R-CNN segmentation in AFOs}
	\begin{table}[h]
		\centering
		\caption{Catch composition estimations for 21 test AFOs using the \textbf{Mask R-CNN} approach for the segmentation and different classifications. We show each AFO's total number of fish, the percentage segmented by our models (evaluation metric for the segmentation, bigger is better), and the catch composition by species percentage (evaluation metric for the classification, similar to GT is better). GT stands for ground truth, S for standard classification, and H for hierarchical.}
		\label{tab:tuna_2_supp_mask}
        \begin{tabularx}{\linewidth}{@{}R|RR|RRR|RRR|RRR|RRR@{}}
            \toprule
			\multicolumn{1}{c|}{\multirow{2}{*}{ID\textsubscript{AFO}}} & \multicolumn{2}{c|}{No. individuals} & \multicolumn{3}{c|}{BET} & \multicolumn{3}{c|}{SKJ} & \multicolumn{3}{c|}{YFT} & \multicolumn{3}{c}{NO\_TARGET} \\ \cmidrule(l){2-15} 
			\multicolumn{1}{c|}{} & \multicolumn{1}{c}{GT} & \multicolumn{1}{c|}{Model} & \multicolumn{1}{c}{GT} & \multicolumn{1}{c}{S} & \multicolumn{1}{c|}{H} & \multicolumn{1}{c}{GT} & \multicolumn{1}{c}{S} & \multicolumn{1}{c|}{H} & \multicolumn{1}{c}{GT} & \multicolumn{1}{c}{S} & \multicolumn{1}{c|}{H} & \multicolumn{1}{c}{GT} & \multicolumn{1}{c}{S} & \multicolumn{1}{c}{H} \\ \midrule
			1\_01 & 327 & 48.3\% & 11.9\% & \textbf{8.2\%} & 6.3\% & 46.2\% & \textbf{58.9\%} & 27.8\% & 39.1\% & \textbf{27.8\%} & 59.5\% & 2.8\% & \textbf{5.1\%} & 6.3\% \\
			1\_02 & 232 & 45.7\% & 1.3\% & \textbf{4.7\%} & 8.5\% & 84.5\% & \textbf{64.2\%} & 38.7\% & 14.2\% & \textbf{30.2\%} & 46.2\% & 0.0\% & \textbf{0.9\%} & 6.6\% \\
			1\_03 & 204 & 46.6\% & 1.0\% & \textbf{2.1\%} & 8.4\% & 90.7\% & \textbf{68.4\%} & 26.3\% & 8.3\% & \textbf{26.3\%} & 58.9\% & 0.0\% & \textbf{3.2\%} & 6.3\% \\
			1\_04 & 239 & 49.4\% & 12.6\% & \textbf{6.8\%} & 5.1\% & 48.1\% & \textbf{39.0\%} & 12.7\% & 39.3\% & \textbf{47.5\%} & 80.5\% & 0.0\% & 6.8\% & \textbf{1.7\%} \\
			1\_05 & 284 & 42.6\% & 12.0\% & 9.1\% & \textbf{9.9\%} & 59.5\% & \textbf{50.4\%} & 25.6\% & 28.5\% & \textbf{38.8\%} & 62.0\% & 0.0\% & \textbf{1.7\%} & 2.5\% \\
			1\_06 & 217 & 44.7\% & 0.0\% & \textbf{3.1\%} & 8.2\% & 87.1\% & \textbf{54.6\%} & 25.8\% & 12.9\% & \textbf{37.1\%} & 62.9\% & 0.0\% & 5.2\% & \textbf{3.1\%} \\
			1\_07 & 185 & 49.7\% & 1.1\% & 2.2\% & \textbf{1.1\%} & 85.4\% & \textbf{47.8\%} & 28.3\% & 13.0\% & \textbf{43.5\%} & 67.4\% & 0.5\% & 6.5\% & \textbf{3.3\%} \\
			1\_08 & 478 & 33.3\% & 2.5\% & 1.9\% & \textbf{2.5\%} & 83.9\% & \textbf{49.1\%} & 30.2\% & 9.4\% & \textbf{40.9\%} & 65.4\% & 4.2\% & 8.2\% & \textbf{1.9\%} \\
			1\_09 & 418 & 40.4\% & 4.1\% & \textbf{4.1\%} & \textbf{4.1\%} & 65.1\% & \textbf{49.1\%} & 29.0\% & 27.5\% & \textbf{43.8\%} & 64.5\% & 3.3\% & \textbf{3.0\%} & 2.4\% \\
			1\_10 & 390 & 43.8\% & 1.3\% & \textbf{0.0\%} & 4.7\% & 87.9\% & \textbf{62.0\%} & 35.1\% & 10.5\% & \textbf{35.7\%} & 56.1\% & 0.3\% & \textbf{2.3\%} & 4.1\% \\
			1\_11 & 455 & 61.5\% & 0.4\% & \textbf{1.1\%} & 3.6\% & 69.0\% & \textbf{57.9\%} & 31.8\% & 26.2\% & \textbf{35.7\%} & 61.8\% & 4.4\% & \textbf{5.4\%} & 2.9\% \\
			1\_12 & 499 & 41.1\% & 4.6\% & 1.0\% & \textbf{2.4\%} & 75.8\% & \textbf{55.6\%} & 27.3\% & 15.0\% & \textbf{35.6\%} & 66.3\% & 4.6\% & 7.8\% & \textbf{3.9\%} \\
			1\_13 & 216 & 53.7\% & 1.4\% & 2.6\% & \textbf{0.9\%} & 50.5\% & \textbf{41.4\%} & 23.3\% & 48.1\% & \textbf{53.4\%} & 74.1\% & 0.0\% & 2.65 & \textbf{1.7\%} \\
			1\_14 & 309 & 48.5\% & 4.2\% & \textbf{2.7\%} & \textbf{2.7\%} & 42.1\% & \textbf{30.0\%} & 12.7\% & 53.7\% & \textbf{62.7\%} & 82.7\% & 0.0\% & 4.7\% & \textbf{2.0\%} \\ \midrule
			2\_01 & 223 & 79.8\% & 2.7\% & \textbf{2.2\%} & 5.6\% & 66.4\% & 89.3\% & \textbf{80.9\%} & 30.9\% & \textbf{8.4\%} & 7.3\% & 0.0\% & \textbf{0.0\%} & 6.2\% \\
			2\_02 & 312 & 57.7\% & 7.1\% & 2.8\% & \textbf{3.9\%} & 85.3\% & 93.3\% & \textbf{83.3\%} & 6.4\% & 2.8\% & \textbf{3.9\%} & 1.3\% & \textbf{1.1\%} & 8.9\% \\
			2\_03 & 344 & 61.9\% & 1.2\% & \textbf{1.4\%} & 3.3\% & 87.8\% & 96.2\% & \textbf{83.6\%} & 10.8\% & 2.3\% & \textbf{5.2\%} & 0.3\% & \textbf{0.0\%} & 8.0\% \\
			2\_04 & 227 & 63.9\% & 4.4\% & 2.8\% & \textbf{4.1\%} & 85.9\% & 93.8\% & \textbf{87.6\%} & 5.7\% & 1.4\% & \textbf{4.1\%} & 4.0\% & 2.1\% & \textbf{4.1\%} \\
			2\_05 & 460 & 46.5\% & 0.2\% & \textbf{0.0\%} & 0.5\% & 77.6\% & 95.3\% & \textbf{87.9\%} & 3.9\% & \textbf{4.2\%} & 5.6\% & 18.3\% & 0.5\% & \textbf{6.1\%} \\
			2\_06 & 232 & 79.7\% & 0.0\% & \textbf{0.0\%} & \textbf{0.0\%} & 92.7\% & 97.8\% & \textbf{90.3\%} & 0.0\% & \textbf{2.2\%} & 4.9\% & 7.3\% & 0.0\% & \textbf{4.9\%} \\
			2\_07 & 402 & 55.0\% & 1.0\% & \textbf{0.0\%} & \textbf{0.0\%} & 75.4\% & 95.9\% & \textbf{83.7\%} & 8.7\% & 3.6\% & \textbf{10.9\%} & 14.9\% & 0.5\% & \textbf{5.4\%} \\ \bottomrule
		\end{tabularx}
	\end{table}
	\vspace*{\fill} 
\end{landscape}
\begin{landscape}
	\vspace*{\fill} 
	\section{YOLOv9 + SAM2 segmentation in AFOs}
	\begin{table}[h]
		\centering
		\caption{Catch composition estimations for 21 test AFOs using the \textbf{YOLOv9 + SAM2} approach for the segmentation and different classifications. We show each AFO's total number of fish, the percentage segmented by our models (evaluation metric for the segmentation, bigger is better), and the catch composition by species percentage (evaluation metric for the classification, similar to GT is better). GT stands for ground truth, S for standard classification, and H for hierarchical.}
		\label{tab:tuna_2_supp_yolo_sam}
		\begin{tabularx}{\linewidth}{@{}R|RR|RRR|RRR|RRR|RRR@{}}
			\toprule
			\multicolumn{1}{c|}{\multirow{2}{*}{ID\textsubscript{AFO}}} & \multicolumn{2}{c|}{No. individuals} & \multicolumn{3}{c|}{BET} & \multicolumn{3}{c|}{SKJ} & \multicolumn{3}{c|}{YFT} & \multicolumn{3}{c}{NO\_TARGET} \\ \cmidrule(l){2-15} 
			\multicolumn{1}{c|}{} & \multicolumn{1}{c}{GT} & \multicolumn{1}{c|}{Model} & \multicolumn{1}{c}{GT} & \multicolumn{1}{c}{S} & \multicolumn{1}{c|}{H} & \multicolumn{1}{c}{GT} & \multicolumn{1}{c}{S} & \multicolumn{1}{c|}{H} & \multicolumn{1}{c}{GT} & \multicolumn{1}{c}{S} & \multicolumn{1}{c|}{H} & \multicolumn{1}{c}{GT} & \multicolumn{1}{c}{S} & \multicolumn{1}{c}{H} \\ \midrule
			1\_01 & 327 & 67.0\% & 11.9\% & \textbf{6.8\%} & 6.4\% & 46.2\% & \textbf{49.8\%} & 23.3\% & 39.1\% & \textbf{39.3\%} & 63.9\% & 2.8\% & \textbf{4.1\%} & 6.4\% \\
			1\_02 & 232 & 60.8\% & 1.3\% & \textbf{5.7\%} & 7.1\% & 84.5\% & \textbf{53.2\%} & 31.2\% & 14.2\% & \textbf{39.7\%} & 56.7\% & 0.0\% & \textbf{1.4\%} & 5.0\% \\
			1\_03 & 204 & 51.0\% & 1.0\% & \textbf{2.9\%} & 7.7\% & 90.7\% & \textbf{61.5\%} & 22.1\% & 8.3\% & \textbf{32.7\%} & 63.5\% & 0.0\% & \textbf{2.9\%} & 6.7\% \\
			1\_04 & 239 & 60.7\% & 12.6\% & \textbf{4.8\%} & 4.1\% & 48.1\% & \textbf{34.5\%} & 11.7\% & 39.3\% & \textbf{52.4\%} & 82.1\% & 0.0\% & 8.3\% & \textbf{2.1\%} \\
			1\_05 & 284 & 58.8\% & 12.0\% & 6.0\% & \textbf{6.6\%} & 59.5\% & \textbf{50.3\%} & 21.6\% & 28.5\% & \textbf{42.5\%} & 69.5\% & 0.0\% & \textbf{1.2\%} & 2.4\% \\
			1\_06 & 217 & 59.9\% & 0.0\% & \textbf{4.6\%} & 7.7\% & 87.1\% & \textbf{49.2\%} & 20.0\% & 12.9\% & \textbf{43.1\%} & 67.7\% & 0.0\% & \textbf{3.1\%} & 4.6\% \\
			1\_07 & 185 & 69.2\% & 1.1\% & \textbf{0.8\%} & 3.9\% & 85.4\% & \textbf{42.2\%} & 25.0\% & 13.0\% & \textbf{53.1\%} & 68.8\% & 0.5\% & 3.9\% & \textbf{2.3\%} \\
			1\_08 & 478 & 46.7\% & 2.5\% & 0.9\% & \textbf{1.3\%} & 83.9\% & \textbf{52.5\%} & 26.5\% & 9.4\% & \textbf{40.8\%} & 70.0\% & 4.2\% & \textbf{5.8\%} & 2.2\% \\
			1\_09 & 418 & 52.2\% & 4.1\% & \textbf{5.0\%} & \textbf{5.0\%} & 65.1\% & \textbf{41.3\%} & 20.6\% & 27.5\% & \textbf{50.9\%} & 72.0\% & 3.3\% & \textbf{2.8\%} & 2.3\% \\
			1\_10 & 390 & 57.7\% & 1.3\% & \textbf{0.9\%} & 3.1\% & 87.9\% & \textbf{55.1\%} & 32.9\% & 10.5\% & \textbf{41.3\%} & 60.4\% & 0.3\% & \textbf{2.7\%} & 3.6\% \\
			1\_11 & 455 & 85.5\% & 0.4\% & \textbf{1.3\%} & 4.9\% & 69.0\% & \textbf{53.0\%} & 28.3\% & 26.2\% & \textbf{38.0\%} & 64.5\% & 4.4\% & 7.7\% & \textbf{2.3\%} \\
			1\_12 & 499 & 47.9\% & 4.6\% & 0.4\% & \textbf{2.9\%} & 75.8\% & \textbf{51.9\%} & 22.6\% & 15.0\% & \textbf{40.6\%} & 70.7\% & 4.6\% & 7.1\% & \textbf{3.8\%} \\
			1\_13 & 216 & 66.7\% & 1.4\% & \textbf{1.4\%} & 2.1\% & 50.5\% & \textbf{38.9\%} & 21.5\% & 48.1\% & \textbf{54.9\%} & 75.0\% & 0.0\% & 4.9\% & \textbf{1.4\%} \\
			1\_14 & 309 & 60.8\% & 4.2\% & 2.7\% & \textbf{3.2\%} & 42.1\% & \textbf{28.7\%} & 12.8\% & 53.7\% & \textbf{66.0\%} & 82.4\% & 0.0\% & 2.7\% & \textbf{1.6\%} \\ \midrule
			2\_01 & 223 & 102.7\% & 2.7\% & \textbf{3.1\%} & 7.0\% & 66.4\% & 89.5\% & \textbf{81.2\%} & 30.9\% & \textbf{7.4\%} & 6.6\% & 0.0\% & \textbf{0.0\%} & 5.2\% \\
			2\_02 & 312 & 84.3\% & 7.1\% & \textbf{2.7\%} & 1.1\% & 85.3\% & 93.9\% & \textbf{84.0\%} & 6.4\% & 2.3\% & \textbf{6.8\%} & 1.3\% & \textbf{1.1\%} & 8.0\% \\
			2\_03 & 344 & 82.3\% & 1.2\% & \textbf{1.1\%} & 3.5\% & 87.8\% & 97.2\% & \textbf{84.1\%} & 10.8\% & 1.4\% & \textbf{3.2\%} & 0.3\% & \textbf{0.4\%} & 9.2\% \\
			2\_04 & 227 & 81.9\% & 4.4\% & 2.2\% & \textbf{3.2\%} & 85.9\% & 93.5\% & \textbf{86.0\%} & 5.7\% & 1.6\% & \textbf{4.3\%} & 4.0\% & \textbf{2.7\%} & 6.5\% \\
			2\_05 & 460 & 67.6\% & 0.2\% & 0.0\% & \textbf{0.3\%} & 77.6\% & 97.1\% & \textbf{83.9\%} & 3.9\% & \textbf{2.6\%} & 6.1\% & 18.3\% & 0.3\% & \textbf{9.6\%} \\
			2\_06 & 232 & 105.2\% & 0.0\% & \textbf{0.0\%} & \textbf{0.0\%} & 92.7\% & 98.8\% & \textbf{91.0\%} & 0.0\% & \textbf{1.2\%} & 2.5\% & 7.3\% & 0.0\% & \textbf{6.6\%} \\
			2\_07 & 402 & 69.9\% & 1.0\% & \textbf{0.4\%} & 0.0\% & 75.4\% & 95.7\% & \textbf{82.2\%} & 8.7\% & 3.2\% & \textbf{7.8\%} & 14.9\% & 0.7\% & \textbf{10.0\%} \\ \bottomrule
		\end{tabularx}
	\end{table}
	\vspace*{\fill} 
\end{landscape}
\section{Expert assigned tuna species}
\begin{xltabular}{\textwidth}{@{}XXXXXXXXXXXX@{}}
	\caption{Species assigned to each of the fishes by the nine experts.}
	\label{tab:tuna_2_supp_experts} \\
	
	\toprule
	\multirow{2}{*}{ID\textsubscript{F}} & \multicolumn{9}{c}{Species assigned by each expert} & \multicolumn{2}{c}{Total} \\
	\cmidrule(l){2-12}
	& 1 & 2 & 3 & 4 & 5 & 6 & 7 & 8 & 9 & YFT & BET \\
	\midrule
	\endfirsthead
	
	\toprule
	\multirow{2}{*}{ID\textsubscript{F}} & \multicolumn{9}{c}{Species assigned by each expert} & \multicolumn{2}{c}{Total} \\
	\cmidrule(l){2-12}
	& 1 & 2 & 3 & 4 & 5 & 6 & 7 & 8 & 9 & YFT & BET \\
	\midrule
	\endhead
	
	\midrule
	\endfoot
	
	\bottomrule
	\endlastfoot
	
	001 & YFT & YFT & YFT & YFT & YFT & YFT & - & - & YFT & 7 & 0 \\
	002 & - & - & YFT & YFT & YFT & YFT & - & - & YFT & 5 & 0 \\
	003 & - & YFT & YFT & YFT & - & YFT & YFT & YFT & YFT & 7 & 0 \\
	004 & - & - & YFT & - & BET & - & - & - & - & 1 & 1 \\
	005 & - & - & YFT & - & YFT & - & - & - & - & 2 & 0 \\
	006 & - & - & - & - & YFT & YFT & - & - & - & 2 & 0 \\
	007 & - & - & - & - & - & - & - & - & BET & 0 & 1 \\
	008 & - & - & - & - & YFT & - & - & - & - & 1 & 0 \\ \midrule
	009 & - & - & - & - & YFT & YFT & - & - & - & 2 & 0 \\
	010 & - & - & - & - & YFT & - & BET & - & - & 1 & 1 \\
	011 & - & - & - & BET & - & - & - & - & BET & 0 & 2 \\
	012 & - & - & - & - & - & - & - & - & YFT & 1 & 0 \\
	013 & - & - & YFT & - & - & BET & - & - & - & 1 & 1 \\
	014 & - & - & - & YFT & YFT & YFT & - & - & - & 3 & 0 \\
	015 & YFT & - & BET & BET & - & BET & - & - & YFT & 2 & 3 \\
	016 & - & YFT & YFT & YFT & YFT & YFT & - & YFT & YFT & 7 & 0 \\
	017 & - & - & - & - & YFT & - & - & - & - & 1 & 0 \\
	018 & - & - & - & - & YFT & - & - & - & - & 1 & 0 \\ \midrule
	019 & - & - & - & BET & - & BET & - & BET & BET & 0 & 4 \\
	020 & YFT & YFT & YFT & YFT & YFT & YFT & - & - & YFT & 7 & 0 \\
	021 & - & - & - & BET & YFT & BET & - & - & - & 1 & 2 \\
	022 & - & - & YFT & BET & YFT & - & BET & - & YFT & 3 & 2 \\
	023 & - & - & - & BET & YFT & YFT & - & - & - & 2 & 1 \\ \midrule
	024 & BET & - & - & BET & BET & BET & BET & - & YFT & 1 & 5 \\
	025 & - & YFT & YFT & YFT & BET & BET & - & YFT & BET & 4 & 3 \\
	026 & - & - & - & - & YFT & - & - & - & - & 1 & 0 \\
	027 & - & - & - & - & YFT & - & - & - & - & 1 & 0 \\ \midrule
	028 & - & - & - & YFT & YFT & BET & - & - & - & 2 & 1 \\
	029 & - & YFT & - & YFT & YFT & - & - & - & - & 3 & 0 \\
	030 & - & - & YFT & YFT & YFT & YFT & - & - & BET & 4 & 1 \\
	031 & - & - & - & - & YFT & - & - & - & BET & 1 & 1 \\
	032 & - & - & - & BET & YFT & BET & - & BET & YFT & 2 & 3 \\
	033 & BET & - & - & BET & - & BET & BET & - & BET & 0 & 5 \\
	034 & - & YFT & YFT & - & YFT & - & - & - & - & 3 & 0 \\ \midrule
	035 & YFT & YFT & YFT & - & YFT & YFT & YFT & - & - & 6 & 0 \\
	036 & - & - & - & - & - & - & - & - & BET & 0 & 1 \\
	037 & - & - & - & YFT & YFT & YFT & - & - & BET & 3 & 1 \\
	038 & - & - & - & - & - & YFT & - & BET & BET & 1 & 2 \\
	039 & - & - & - & BET & - & - & - & - & - & 0 & 1 \\
	040 & - & - & - & - & YFT & - & - & - & - & 1 & 0 \\ \midrule
	041 & - & - & - & - & YFT & YFT & - & - & - & 2 & 0 \\
	042 & - & - & - & - & BET & - & - & - & BET & 0 & 2 \\
	043 & - & - & - & YFT & BET & - & - & - & YFT & 2 & 1 \\
	044 & - & - & - & YFT & - & YFT & - & - & - & 2 & 0 \\
	045 & YFT & YFT & YFT & YFT & YFT & YFT & YFT & YFT & YFT & 9 & 0 \\
	046 & - & - & - & - & YFT & YFT & - & - & BET & 2 & 1 \\
	047 & - & - & - & - & YFT & - & - & - & YFT & 2 & 0 \\
	048 & - & - & - & - & - & YFT & - & - & - & 1 & 0 \\
	049 & - & - & - & - & YFT & - & - & - & YFT & 2 & 0 \\
	050 & - & - & - & - & YFT & - & - & - & - & 1 & 0 \\ \midrule
	051 & - & - & - & YFT & - & YFT & - & - & YFT & 3 & 0 \\
	052 & - & - & YFT & YFT & YFT & - & - & - & - & 3 & 0 \\
	053 & - & - & YFT & YFT & YFT & YFT & - & - & BET & 4 & 1 \\
	054 & - & - & - & - & - & YFT & - & - & BET & 1 & 1 \\
	055 & - & YFT & - & BET & - & - & - & - & - & 1 & 1 \\
	056 & - & - & - & YFT & - & YFT & - & - & BET & 2 & 1 \\
	057 & - & - & YFT & YFT & YFT & YFT & YFT & - & YFT & 6 & 0 \\
	058 & - & - & - & YFT & - & - & - & - & - & 1 & 0 \\
	059 & - & - & - & - & BET & - & - & - & YFT & 1 & 1 \\
	060 & - & - & - & - & - & - & - & YFT & - & 1 & 0 \\
	061 & - & - & - & - & - & - & - & - & BET & 0 & 1 \\ \midrule
	062 & - & - & - & - & - & - & - & - & BET & 0 & 1 \\
	063 & - & - & - & BET & - & - & - & - & - & 0 & 1 \\
	064 & - & - & - & - & YFT & - & - & BET & BET & 1 & 2 \\
	065 & BET & - & - & BET & YFT & YFT & - & - & YFT & 3 & 2 \\
	066 & - & - & YFT & YFT & YFT & YFT & YFT & - & YFT & 6 & 0 \\
	067 & - & - & - & - & - & - & - & - & YFT & 1 & 0 \\
	068 & - & - & - & - & YFT & YFT & - & - & BET & 2 & 1 \\
	069 & - & - & - & - & YFT & - & - & - & - & 1 & 0 \\ \midrule
	070 & - & - & - & BET & BET & YFT & - & - & YFT & 2 & 2 \\
	071 & - & - & - & YFT & YFT & - & - & - & YFT & 3 & 0 \\
	072 & BET & - & BET & BET & BET & YFT & BET & BET & YFT & 2 & 6 \\
	073 & - & - & - & YFT & YFT & - & - & - & - & 2 & 0 \\
	074 & - & - & - & - & YFT & - & - & - & - & 1 & 0 \\ \midrule
	075 & - & YFT & YFT & YFT & YFT & YFT & - & - & YFT & 6 & 0 \\
	076 & BET & - & - & YFT & YFT & YFT & - & BET & YFT & 4 & 2 \\
	077 & - & - & - & YFT & - & - & - & - & YFT & 2 & 0 \\
	078 & - & - & YFT & YFT & - & - & BET & - & YFT & 3 & 1 \\
	079 & - & - & - & - & YFT & YFT & - & - & - & 2 & 0 \\
	080 & - & - & - & - & YFT & - & - & - & YFT & 2 & 0 \\ \midrule
	081 & - & - & - & - & YFT & - & - & - & YFT & 2 & 0 \\
	082 & - & - & - & - & - & - & - & - & YFT & 1 & 0 \\
	083 & - & - & - & - & - & - & - & - & BET & 0 & 1 \\
	084 & - & - & - & YFT & BET & - & - & - & BET & 1 & 2 \\
	085 & BET & - & - & YFT & YFT & - & - & - & BET & 2 & 2 \\
	086 & BET & - & - & YFT & - & YFT & - & BET & BET & 2 & 3 \\
	087 & BET & - & BET & YFT & - & YFT & BET & - & BET & 2 & 4 \\
	088 & - & - & - & YFT & YFT & - & - & - & BET & 2 & 1 \\
	089 & - & - & - & - & BET & - & - & - & - & 0 & 1 \\ \midrule
	090 & - & - & - & - & YFT & YFT & - & - & - & 2 & 0 \\
	091 & - & - & - & BET & - & - & - & - & YFT & 1 & 1 \\
	092 & BET & - & - & - & - & - & - & - & YFT & 1 & 1 \\
	093 & - & - & - & YFT & - & YFT & - & BET & BET & 2 & 2 \\
	094 & - & - & - & YFT & - & - & - & - & BET & 1 & 1 \\
	095 & - & - & BET & - & BET & - & BET & - & BET & 0 & 4 \\
	096 & - & YFT & - & YFT & - & - & - & - & BET & 2 & 1 \\
	097 & - & - & - & - & - & - & - & - & BET & 0 & 1 \\
	098 & - & - & - & - & BET & - & - & - & - & 0 & 1 \\
	099 & - & - & - & - & YFT & - & - & - & - & 1 & 0 \\
	100 & - & - & - & - & BET & - & - & - & - & 0 & 1 \\
	101 & - & - & - & - & YFT & - & - & - & - & 1 & 0 \\
	102 & - & - & - & - & BET & - & - & - & - & 0 & 1 \\
	103 & - & - & - & - & YFT & - & - & - & - & 1 & 0 \\ \midrule
	104 & BET & BET & BET & BET & BET & BET & BET & BET & BET & 0 & 9 \\
	105 & - & - & BET & BET & - & - & - & - & BET & 0 & 3 \\
	106 & - & - & - & - & - & - & - & - & BET & 0 & 1 \\ \midrule
	107 & - & - & - & - & YFT & YFT & - & - & - & 2 & 0 \\
	108 & - & - & - & YFT & YFT & - & - & - & YFT & 3 & 0 \\
	109 & - & - & - & BET & BET & - & - & - & YFT & 1 & 2 \\
	110 & - & YFT & YFT & YFT & YFT & YFT & BET & - & YFT & 6 & 1 \\
	111 & - & - & - & YFT & - & - & - & YFT & YFT & 3 & 0 \\
	112 & - & - & - & YFT & - & - & - & - & YFT & 2 & 0 \\
	113 & - & - & - & - & YFT & YFT & - & - & YFT & 3 & 0 \\
	114 & - & - & - & - & BET & - & - & - & - & 0 & 1 \\
	115 & - & - & - & - & YFT & - & - & - & - & 1 & 0 \\ \midrule
	116 & - & - & - & YFT & YFT & YFT & YFT & - & YFT & 5 & 0 \\
	117 & - & - & - & - & YFT & - & - & - & YFT & 2 & 0 \\
	118 & - & - & - & - & YFT & - & - & - & YFT & 2 & 0 \\
	119 & - & - & - & - & - & - & - & BET & - & 0 & 1 \\
	120 & - & - & - & YFT & YFT & YFT & - & - & - & 3 & 0 \\
	121 & - & - & YFT & YFT & YFT & YFT & - & - & YFT & 5 & 0 \\
	122 & - & - & - & - & - & - & - & - & BET & 0 & 1 \\
	123 & - & - & - & - & YFT & - & - & - & BET & 1 & 1 \\
	124 & - & YFT & - & - & YFT & - & - & - & - & 2 & 0 \\
	125 & - & - & - & - & YFT & - & - & - & - & 1 & 0 \\ \midrule
	126 & - & - & - & - & - & - & - & - & BET & 0 & 1 \\
	127 & - & - & - & - & YFT & - & - & - & BET & 1 & 1 \\
	128 & - & - & - & YFT & YFT & - & - & - & BET & 2 & 1 \\
	129 & - & - & - & YFT & BET & - & - & - & BET & 1 & 2 \\
	130 & - & - & - & - & - & - & - & - & BET & 0 & 1 \\
	131 & - & - & - & YFT & YFT & - & - & - & BET & 2 & 1 \\
	132 & - & - & - & YFT & - & - & - & - & - & 1 & 0 \\
	133 & - & - & - & YFT & - & - & - & - & BET & 1 & 1 \\
	134 & - & - & - & YFT & YFT & - & - & - & BET & 2 & 1 \\
	135 & BET & - & BET & BET & YFT & YFT & BET & BET & BET & 2 & 6 \\ \midrule
	136 & YFT & YFT & YFT & YFT & YFT & - & - & - & BET & 5 & 1 \\
	137 & - & YFT & - & YFT & BET & - & - & - & BET & 2 & 2 \\
	138 & - & - & - & BET & BET & - & BET & BET & BET & 0 & 5 \\
	139 & - & - & - & - & YFT & - & - & - & - & 1 & 0 \\ \midrule
	140 & BET & BET & - & BET & BET & - & BET & BET & YFT & 1 & 6 \\
	141 & - & - & - & - & - & - & - & - & YFT & 1 & 0 \\
	142 & - & - & - & - & BET & - & - & - & YFT & 1 & 1 \\
	143 & - & - & - & - & BET & - & - & - & YFT & 1 & 1 \\ \midrule
	144 & - & - & - & YFT & - & - & - & - & BET & 1 & 1 \\
	145 & - & - & - & YFT & - & - & - & - & BET & 1 & 1 \\
	146 & - & - & - & YFT & YFT & - & YFT & YFT & BET & 4 & 1 \\
	147 & - & - & - & - & YFT & - & - & - & - & 1 & 0 \\
	148 & - & - & - & - & BET & - & - & - & - & 0 & 1 \\ \midrule
	149 & YFT & YFT & YFT & - & YFT & - & YFT & YFT & YFT & 7 & 0 \\
	150 & - & - & - & - & - & - & - & - & BET & 0 & 1 \\
	151 & - & - & - & - & - & - & - & - & BET & 0 & 1 \\ \midrule
	152 & BET & BET & YFT & BET & BET & - & BET & BET & BET & 1 & 7 \\
	153 & - & - & - & YFT & - & - & - & - & - & 1 & 0 \\
	154 & - & - & - & - & - & - & - & - & BET & 0 & 1 \\
	155 & - & - & - & - & BET & - & - & - & - & 0 & 1 \\ \midrule
	156 & - & BET & - & YFT & YFT & - & BET & YFT & YFT & 4 & 2 \\
	157 & - & - & - & YFT & - & - & - & - & - & 1 & 0 \\
	158 & - & - & - & YFT & - & - & - & - & BET & 1 & 1 \\ \midrule
	159 & BET & BET & - & BET & BET & - & - & - & YFT & 1 & 4 \\
	160 & - & - & - & YFT & YFT & - & YFT & YFT & YFT & 5 & 0 \\
	161 & - & YFT & - & YFT & BET & - & - & - & YFT & 3 & 1 \\
	162 & - & - & - & YFT & YFT & - & - & - & BET & 2 & 1 \\
	163 & - & - & - & - & - & - & - & - & YFT & 1 & 0 \\
	164 & - & - & - & - & - & - & - & - & YFT & 1 & 0 \\ \midrule
	165 & - & - & - & YFT & BET & - & BET & - & BET & 1 & 3 \\
	166 & - & - & YFT & YFT & YFT & YFT & - & BET & BET & 4 & 2 \\ \midrule
	167 & YFT & YFT & YFT & YFT & YFT & YFT & BET & YFT & BET & 7 & 2 \\
	168 & - & - & - & YFT & - & - & - & - & - & 1 & 0 \\
	169 & - & - & - & - & - & - & - & - & YFT & 1 & 0 \\ \midrule
	170 & - & - & - & YFT & YFT & - & YFT & YFT & YFT & 5 & 0 \\
	171 & YFT & - & - & - & YFT & - & - & - & YFT & 3 & 0 \\ \midrule
	172 & BET & - & - & BET & - & - & BET & BET & BET & 0 & 5 \\
	173 & - & - & - & YFT & - & - & - & - & - & 1 & 0 \\
	174 & - & - & - & - & YFT & - & - & - & YFT & 2 & 0 \\
	175 & - & - & - & - & YFT & - & - & - & - & 1 & 0 \\ \midrule
	176 & - & - & YFT & YFT & BET & YFT & YFT & BET & YFT & 5 & 2 \\
	177 & BET & - & - & YFT & YFT & - & - & - & - & 2 & 1 \\
	178 & - & - & - & BET & - & - & - & - & - & 0 & 1 \\
	179 & - & - & - & YFT & - & - & - & - & - & 1 & 0 \\ \midrule
	180 & - & - & - & BET & - & - & - & - & BET & 0 & 2 \\
	181 & - & BET & - & BET & - & - & - & - & BET & 0 & 3 \\
	182 & - & - & BET & BET & BET & - & - & - & BET & 0 & 4 \\
	183 & YFT & BET & BET & BET & BET & YFT & YFT & BET & BET & 3 & 6 \\
	184 & - & YFT & - & BET & YFT & - & - & - & BET & 2 & 2 \\
	185 & - & - & - & BET & - & - & - & - & BET & 0 & 2 \\
	186 & - & - & - & - & YFT & - & - & - & YFT & 2 & 0 \\ \midrule
	187 & - & - & - & BET & YFT & YFT & - & - & YFT & 3 & 1 \\
	188 & - & - & - & BET & BET & - & - & - & - & 0 & 2 \\
	189 & YFT & BET & BET & BET & BET & - & YFT & BET & BET & 2 & 6 \\
	190 & - & - & YFT & YFT & BET & - & - & - & YFT & 3 & 1 \\
	191 & - & - & - & YFT & - & - & - & - & YFT & 2 & 0 \\
	192 & - & - & - & BET & - & - & - & - & BET & 0 & 2 \\
	193 & - & - & - & YFT & YFT & - & - & - & - & 2 & 0 \\
	194 & - & - & - & - & - & - & - & - & YFT & 1 & 0 \\
	195 & - & - & - & - & - & - & - & - & YFT & 1 & 0 \\
	196 & - & - & - & - & - & - & - & - & BET & 0 & 1 \\ \midrule
	197 & - & - & BET & YFT & BET & - & - & - & BET & 1 & 3 \\
	198 & - & - & - & BET & BET & - & BET & - & BET & 0 & 4 \\
	199 & BET & BET & - & BET & BET & - & - & BET & BET & 0 & 6 \\
	200 & BET & YFT & - & BET & BET & - & - & - & BET & 1 & 4 \\
	201 & - & - & - & YFT & BET & - & - & - & BET & 1 & 2 \\
	202 & - & - & - & YFT & YFT & - & - & - & YFT & 3 & 0 \\
	203 & - & - & BET & YFT & BET & - & - & - & BET & 1 & 3 \\
	204 & - & - & - & YFT & - & - & - & - & BET & 1 & 1 \\
	205 & - & - & - & YFT & BET & - & - & - & BET & 1 & 2 \\
	206 & - & - & - & YFT & - & - & - & - & YFT & 2 & 0 \\
	207 & - & - & - & - & - & - & - & - & BET & 0 & 1 \\
	208 & - & - & - & - & - & - & - & - & BET & 0 & 1 \\ \midrule
	209 & - & - & - & - & BET & - & - & - & BET & 0 & 2 \\
	210 & - & - & YFT & YFT & BET & - & BET & BET & YFT & 3 & 3 \\
	211 & - & - & - & YFT & BET & - & - & - & YFT & 2 & 1 \\
	212 & - & - & - & YFT & - & - & - & - & - & 1 & 0 \\
	213 & - & - & - & - & YFT & - & - & - & - & 1 & 0 \\
	214 & - & - & - & - & YFT & - & - & - & - & 1 & 0 \\ \midrule
	215 & - & - & - & YFT & BET & - & - & - & BET & 1 & 2 \\
	216 & - & BET & BET & BET & BET & - & BET & - & BET & 0 & 6 \\
	217 & - & - & - & BET & BET & - & - & BET & YFT & 1 & 3 \\
	218 & - & - & - & BET & BET & - & - & - & BET & 0 & 3 \\
	219 & - & - & - & - & - & - & - & - & BET & 0 & 1 \\
	220 & - & - & - & - & - & - & - & - & YFT & 1 & 0 \\
	221 & - & - & - & - & BET & - & - & - & - & 0 & 1 \\ \midrule
	222 & - & YFT & BET & YFT & YFT & YFT & - & YFT & YFT & 6 & 1 \\
	223 & BET & - & - & YFT & BET & - & - & - & YFT & 2 & 2 \\
	224 & - & - & - & BET & - & - & - & - & - & 0 & 1 \\
	225 & - & - & - & YFT & YFT & - & - & - & YFT & 3 & 0 \\
	226 & - & - & BET & YFT & BET & - & BET & - & YFT & 2 & 3 \\
	227 & - & BET & - & BET & - & - & - & - & BET & 0 & 3 \\
	228 & - & - & - & YFT & YFT & - & - & - & - & 2 & 0 \\
	229 & - & - & - & - & - & - & - & - & YFT & 1 & 0 \\
	230 & - & - & - & - & - & - & - & - & YFT & 1 & 0 \\
	231 & - & - & - & - & YFT & - & - & - & - & 1 & 0 \\ \midrule
	232 & - & - & - & - & - & - & BET & - & YFT & 1 & 1 \\
	233 & - & - & - & YFT & - & - & - & - & YFT & 2 & 0 \\
	234 & - & - & - & YFT & - & - & - & - & YFT & 2 & 0 \\
	235 & - & - & - & BET & YFT & - & - & - & YFT & 2 & 1 \\
	236 & - & - & - & YFT & BET & - & - & BET & YFT & 2 & 2 \\
	237 & BET & - & - & - & - & - & - & - & - & 0 & 1 \\
	238 & - & - & - & - & - & - & - & - & BET & 0 & 1 \\ \midrule
	239 & YFT & YFT & YFT & YFT & YFT & - & YFT & YFT & YFT & 8 & 0 \\
	240 & - & - & - & YFT & - & - & - & - & YFT & 2 & 0 \\
	241 & - & - & - & - & - & - & - & - & YFT & 1 & 0 \\ \midrule
	242 & YFT & - & YFT & YFT & YFT & - & YFT & YFT & YFT & 7 & 0 \\
	243 & - & - & - & YFT & - & - & - & - & - & 1 & 0 \\
	244 & - & BET & - & - & BET & - & - & - & YFT & 1 & 2 \\
	245 & - & - & - & YFT & - & - & - & - & YFT & 2 & 0 \\
	246 & - & - & - & YFT & YFT & - & - & - & YFT & 3 & 0 \\
	247 & - & - & - & - & YFT & - & - & - & - & 1 & 0 \\ \midrule
	248 & - & - & - & - & - & - & - & - & YFT & 1 & 0 \\
	249 & - & - & - & - & - & - & - & - & BET & 0 & 1 \\
	250 & - & - & YFT & YFT & YFT & - & - & - & YFT & 4 & 0 \\
	251 & - & - & - & - & - & - & - & - & YFT & 1 & 0 \\
	252 & - & - & - & - & - & - & BET & BET & BET & 0 & 3 \\
	253 & YFT & - & - & YFT & - & - & - & - & YFT & 3 & 0 \\
	254 & - & YFT & - & - & YFT & - & - & - & YFT & 3 & 0 \\
	255 & - & - & - & YFT & YFT & - & - & - & YFT & 3 & 0 \\ \midrule
	256 & YFT & YFT & YFT & - & YFT & YFT & YFT & YFT & YFT & 8 & 0 \\
	257 & - & YFT & YFT & - & YFT & - & - & - & YFT & 4 & 0 \\
\end{xltabular}

\end{document}